\documentclass[10pt,twocolumn,letterpaper]{article}

\usepackage[pagenumbers]{cvpr} 
\usepackage{booktabs}
\usepackage{enumitem}
\usepackage{multirow}
\usepackage{wrapfig}
\usepackage{colortbl}
\usepackage[normalem]{ulem}
\usepackage{microtype}
\usepackage{comment}
\usepackage[hang,flushmargin]{footmisc}

\NewDocumentCommand\emojipaint{}{
    \includegraphics[scale=0.095]{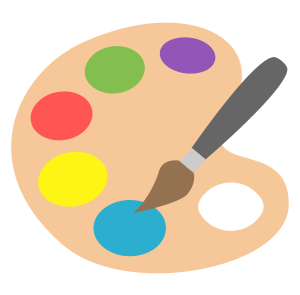}
}

\setlist[itemize]{align=parleft,left=0pt,topsep=1mm,itemsep=0mm,parsep=1mm}

\definecolor{azure(colorwheel)}{rgb}{0.0, 0.5, 1.0}
\definecolor{nicegreen}{rgb}{0.0, 0.7, 0.1}
\definecolor{yw}{rgb}{0.01176, 0.5490, 0.5490}
\definecolor{ashblue}{rgb}{0.36, 0.54, 0.66}
\definecolor{ashgrey}{rgb}{0.7, 0.75, 0.71}
\definecolor{applegreen}{rgb}{0.55, 0.71, 0.0}
\definecolor{blue}{rgb}{0.0, 0.0, 1.0}
\definecolor{postechred}{rgb}{0.784, 0.003, 0.313}
\definecolor{ywg}{rgb}{0.9960, 0.8984, 0.5859}
\definecolor{ballblue}{rgb}{0.13, 0.67, 0.8}
\definecolor{cornellred}{rgb}{0.7, 0.11, 0.11}
\definecolor{darkcyan}{rgb}{0.0, 0.55, 0.55}
\definecolor{CuGray}{gray}{0.9}
\definecolor{airforceblue}{rgb}{0.36, 0.54, 0.66}
\definecolor{rev}{rgb}{0.784, 0.003, 0.313}
\definecolor{pink}{cmyk}{0, 0.7808, 0.4429, 0.1412}
\definecolor{amethyst}{rgb}{0.6, 0.4, 0.8}
\definecolor{black}{rgb}{0.0, 0.0, 0.0}
\definecolor{tb3_yellow}{rgb}{0.996, 1.0, 0.6}
\definecolor{tb3_orange}{rgb}{0.980, 0.8, 0.604}
\definecolor{tb3_red}{rgb}{0.972, 0.6, 0.6}
\definecolor{dimgray}{rgb}{0.41, 0.41, 0.41}
\definecolor{brickred}{rgb}{0.8, 0.25, 0.33}
\definecolor{bleudefrance}{rgb}{0.19, 0.55, 0.91}
\definecolor{blue(ncs)}{rgb}{0.265, 0.445, 0.765}
\definecolor{blue(ryb)}{rgb}{0.01, 0.28, 1.0}
\definecolor{orange}{rgb}{1.0, 0.49, 0.0}
\definecolor{Gray}{gray}{0.88}
\definecolor{green(ncs)}{rgb}{0.0, 0.62, 0.42}
\definecolor{brightpink}{rgb}{1.0, 0.0, 0.5}

\definecolor{kellygreen}{rgb}{0.3, 0.73, 0.09}
\newcommand{\colorref}[1]{{\color{cornellred}{#1}}}

\newcolumntype{g}{>{\columncolor{CuGray}}c}
\newcolumntype{z}{>{\columncolor{CuGray}}l}

\renewcommand{\paragraph}[1]{\vspace{1mm}\noindent\textbf{#1.}\,\,}

\usepackage{xspace}

\makeatletter
\def\@fnsymbol#1{\ensuremath{\ifcase#1\or *\or \dagger\or \ddagger\or
   \mathsection\or \mathparagraph\or \|\or **\or \dagger\dagger
   \or \ddagger\ddagger \else\@ctrerr\fi}}
\makeatother

\def\onedot{.\@\xspace}
\def\eg{\emph{e.g}\onedot} 
\def\ie{\emph{i.e}\onedot}

\def\etal{\emph{et al}\onedot}
\def\ours{\emph{Paint-it}}
\def\dcpbr{\text{DC-PBR}}

\newcommand{\Sref}[1]{Sec.~\ref{#1}}
\newcommand{\Eref}[1]{Eq.~(\ref{#1})}
\newcommand{\Fref}[1]{Fig.~\ref{#1}}
\newcommand{\Tref}[1]{Table~\ref{#1}}

\newcommand{\bk}{{\mathbf{k}}}

\newcommand{\bn}{{\mathbf{n}}}

\newcommand{\bp}{{\mathbf{p}}}

\newcommand{\bx}{{\mathbf{x}}}

\newcommand{\bz}{{\mathbf{z}}}

\newcommand{\bG}{\mathbf{G}}

\newcommand{\bI}{\mathbf{I}}

\newcommand{\bK}{\mathbf{K}}

\newcommand{\bM}{\mathbf{M}}

\newcommand{\bomega}{\mbox{\boldmath $\omega$}}

\newcommand{\Real}{\mathbb R}

\newcommand{\be}{\begin{eqnarray}}
\newcommand{\ee}{\end{eqnarray}}
\newcommand{\bee}{\begin{eqnarray*}}
\newcommand{\eee}{\end{eqnarray*}}

\newcommand{\matrixb}{\left[ \begin{array}}
\newcommand{\matrixe}{\end{array} \right]}

\newcommand{\argmin}{\operatornamewithlimits{\arg \min}}

\usepackage[hang,flushmargin]{footmisc}
\usepackage{microtype}

\definecolor{cvprblue}{rgb}{0.21,0.49,0.74}
\usepackage[pagebackref,breaklinks,colorlinks,citecolor=cvprblue,linkcolor=cornellred]{hyperref}

\def\paperID{3692} 
\def\confName{CVPR}
\def\confYear{2024}

\title{\emojipaint $\ours$: 
Text-to-Texture Synthesis via Deep Convolutional\\ Texture Map Optimization and Physically-Based Rendering
}

\def\authorBlock{
    Kim Youwang${}^{1,2,4*}$ \qquad
    Tae-Hyun Oh${}^{4,5,6}$ \qquad
    Gerard Pons-Moll${}^{1,2,3}$\vspace{3mm} \\
   \small{${}^{1}$University of T\"ubingen\quad ${}^{2}$T\"ubingen AI Center, Germany\quad ${}^{3}$Max Planck Institute for Informatics, Germany}\\
   \small{${}^{4}$Dept. of Electrical Engineering, POSTECH\quad${}^{5}$Grad. School of AI, POSTECH}\\
   \small{${}^{6}$Institute for Convergence Research and Education in Advanced Technology, Yonsei University}\\ \vspace{-5mm}
}
\author{\authorBlock}

\begin{document}
\twocolumn[{%
\renewcommand\twocolumn[1][]{#1}%
\maketitle 
\vspace{-3mm}
\begin{center}
\vspace{-5mm}
  \centering
  \captionsetup{type=figure}
  \includegraphics[width=\linewidth]{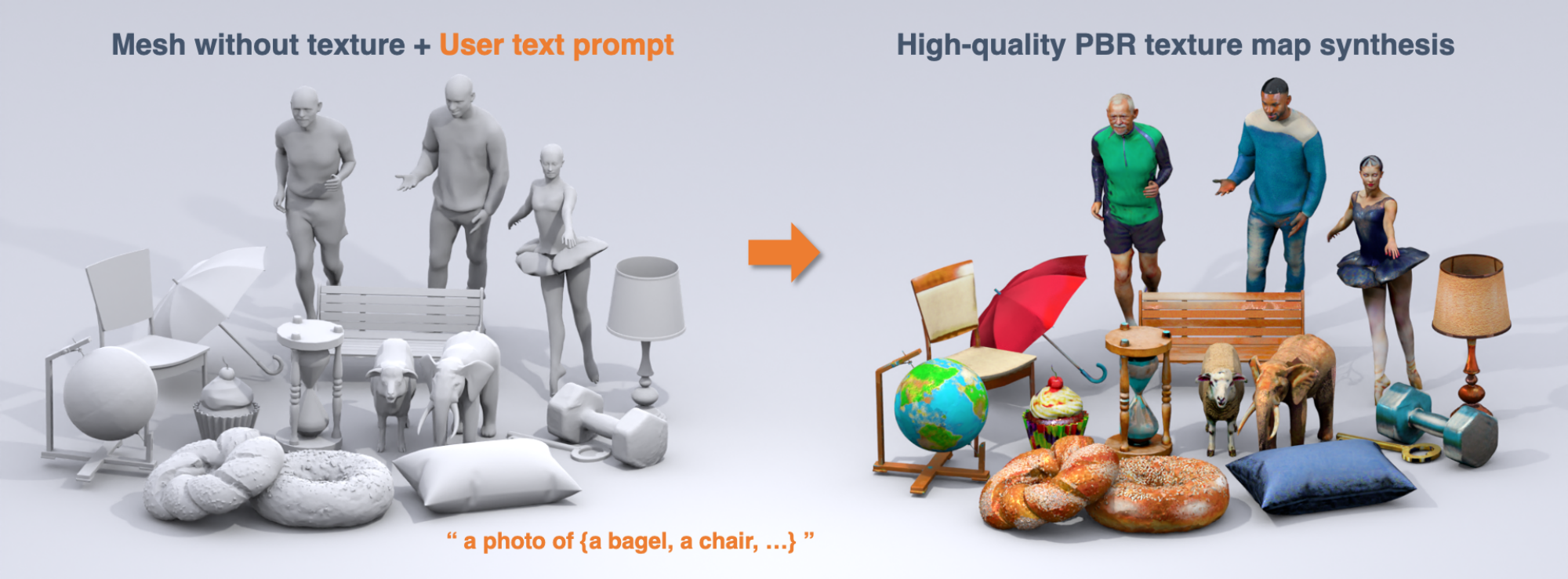}\vspace{-2mm}
   \captionof{figure}{\textbf{$\ours$.} Given an untextured 3D mesh and the text description describing the desired appearance of the 3D mesh, $\ours$ automatically synthesizes high-fidelity physically-based rendering (PBR) texture maps by neural re-parameterized texture map optimization.
   }
   \vspace{1.5mm}
   \label{fig:teaser}
\end{center}
}]
{
  \renewcommand{\thefootnote}%
    {\fnsymbol{footnote}}
  \footnotetext[1]{Work done during a visiting research period at the University of T\"ubingen.}
}

\begin{abstract}\vspace{-3mm}
We present Paint-it, a text-driven high-fidelity texture map synthesis method for 3D meshes via neural re-parameterized texture optimization.
Paint-it synthesizes texture maps from a text description by synthesis-through-optimization, exploiting the Score-Distillation Sampling (SDS).
We observe that directly applying SDS yields undesirable texture quality due to its noisy gradients.
We reveal the importance of 
texture parameterization when using SDS.
Specifically, we propose Deep Convolutional Physically-Based Rendering (DC-PBR) parameterization, 
which re-parameterizes the physically-based rendering (PBR) texture maps with 
randomly initialized 
convolution-based neural kernels, instead of a standard pixel-based parameterization.
We show that DC-PBR
inherently schedules the optimization curriculum according to texture frequency
and naturally filters out the noisy signals from SDS.
In experiments, 
Paint-it obtains
remarkable quality PBR texture maps within 15 min., given only a text description.
We demonstrate the generalizability and practicality of Paint-it by synthesizing high-quality texture maps for large-scale mesh datasets and showing test-time applications such as relighting and material control using a popular graphics engine.
Project page:
\href{https://kim-youwang.github.io/paint-it}{https://kim-youwang.github.io/paint-it}.
\end{abstract}
\vspace{-1.5mm}


\begin{figure*}[t]
\centering
\includegraphics[width=\linewidth]{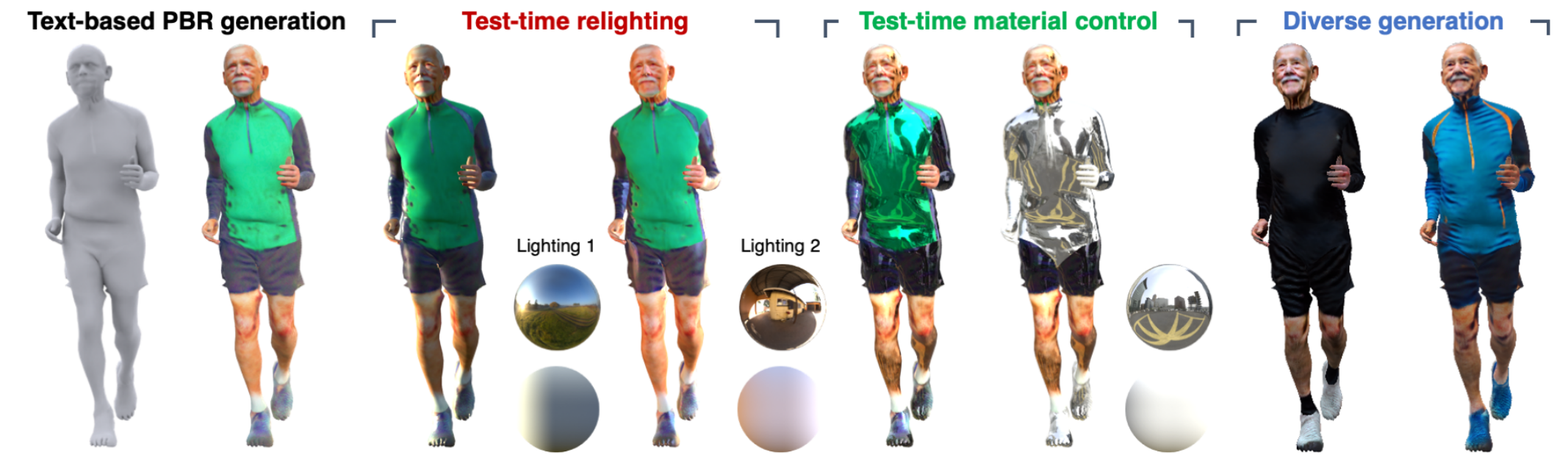}
   \caption{\textbf{$\ours$: Practical applications.}
   Using the synthesized PBR texture maps
   of $\ours$ 
   and commercial graphics engines, \eg, Blender, 
   we can (1) relight the mesh by changing High-Dynamic Range (HDR) environmental lighting (see the balls) and (2) control the material properties at test-time.
   We can also simulate
   diverse appearance by synthesizing different PBR texture maps for the same mesh.
   }
\label{fig:relight}
\end{figure*}

\vspace{-6mm}
\section{Introduction}
\label{sec:intro}

Crafting realistic and diverse 3D assets is the key component in industrial fields such as movies, games, and AR/VR applications.
Professional graphic designers strive to create realistic or creative virtual humans, animals, and objects. 
Still, the hand-designed generation of realistically textured 3D
objects 
requires cumbersome and time-consuming efforts with intensive labor and the pain of creation.

To reduce such burdens, methods for generating 
diverse 3D assets
have been extensively studied~\cite{park2019deepsdf,jiajun20163dgan,chen2022gdna,ma20autoenclother,pavlakos2019smplx,chan2022eg3d,dong2023ag3d}.
Notably, the recent progress in neural volumetric representations, \eg, NeRF~\cite{mildenhall2020nerf} and 
diffusion models~\cite{rombach2021highresolution,saharia2022imagen}
have advanced the development of text-driven 3D asset generation~\cite{jiang2023avatarcraft,poole2022dreamfusion,huang2023dreamwaltz,lin2023magic3d}, which leverages a cheaper guidance, \ie, text description.
%
While these methods 
generate coarse geometries and textures, the 
qualities are still unsatisfactory.
Moreover, to use the generated assets in real graphics engines, \eg, Blender, one must convert the implicit volumetric geometries and textures into the explicit mesh and compatible texture maps, which makes them impractical.
%
Manual extraction of mesh surfaces and unwrapping of textures could be performed, but it still has limitations.
%
The unwrapped texture maps inevitably have heterogeneous texture mappings, so we cannot easily transfer or edit them, which is crucial for generating diverse 3D assets.

Recently, a line of work~\cite{chen2023text2tex,metzer2022latent,richardson2023texture,cao2023texfusion} has been tackling the task of text-driven texture synthesis for practical use.
Instead of generating an entire geometry and texture from scratch, the task aims at synthesizing diverse texture maps on top of the given mesh, conditioned on a text description. 
While many 3D meshes can be reused in the practical production pipeline, the texture maps should be diverse.
For example, a single car mesh can be repeatedly used for making a game, but artists should create distinct texture maps to model different appearances. 
In this vein, text-driven texture map synthesis tries to revolutionize the current repetitive and exhaustive appearance modeling pipeline.
However, existing methods~\cite{chen2023text2tex,richardson2023texture,cao2023texfusion} are limited in that they first conditionally generate latent or RGB images and back-project the colors onto the mesh. 
Although the back-projected colors may look plausible,
they may introduce implausible textures since the 
back-projection cannot model material properties or the complicated reflections on the 3D surfaces.

To address these difficulties, we propose $\ours$, which 
synthesizes high-fidelity texture maps,
given a mesh without texture and the text description via synthesis-through-optimization.
The main contribution of our work is the analysis and investigation of a proper texture representation, which allows easier optimization with the Score-Distillation Sampling (SDS)~\cite{poole2022dreamfusion}.
When optimizing the texture maps, we introduce $\dcpbr$, the 
Deep Convolutional Physically-Based Rendering re-parameterization.
We optimize the
neural surrogate 
of the physically-based rendering (PBR) texture maps rather than directly optimizing the pixel values of the texture maps.
$\dcpbr$ formulates coupled optimization variables with diverse frequencies and serves as an implicit texture prior.
%
In our analysis, we show that the $\dcpbr$, which uses randomly initialized convolution-based neural kernels, naturally imposes the frequency-scheduled learning, which helps filter out high-frequency noisy SDS signals during the optimization.
Furthermore, since $\dcpbr$ re-parameterizes the disentangled texture maps; diffuse, roughness \& metalness, and normal maps,
we simulate physical properties such as the bidirectional reflectance distribution function (BRDF), yielding photorealistic synthesis results.
Overall, we observe that the proposed DC-PBR better interacts with the SDS loss than the diffuse-only texture representation.  
In experiments, we demonstrate that $\ours$ produces
high-quality texture maps for general 3D meshes, \eg, objects, humans, and animals (\Fref{fig:teaser}). 
Also, $\ours$ synthesizes a remarkable quality of texture maps compared to competing methods.
As a favorable byproduct, the synthesized PBR
texture maps are compatible with
the popular graphics engine and can be seamlessly integrated into relighting and material control pipelines (\Fref{fig:relight}).
We summarize our main contributions as follows.
\begin{itemize}
    \item We propose $\ours$, a text-driven synthesis of high-fidelity PBR texture maps, which support practical test-time applications compatible with graphics engines.
    \item We identify the difficulties of synthesizing PBR texture maps in pixel-based parameterization.
    \item We introduce $\dcpbr$, a deep convolutional PBR texture map re-parameterization, and empirical analysis of $\dcpbr$'s benefit when optimizing with the noisy signal, \eg, Score-Distillation Sampling (SDS).
\end{itemize}

\section{Related Work} 
\label{sec:related}
%
%

\paragraph{Text-driven 3D Asset Generation}
Recently, a few impressive works have proposed remarkable 3D asset generation methods that require only simple text prompts~\cite{poole2022dreamfusion,lin2023magic3d,youwang2022clipactor,hong2022avatarclip,jiang2023avatarcraft,huang2023dreamwaltz,jain2021dreamfields,oscar2022text2mesh,chen2023text2tex,chen2023fantasia,chen2022tango}.
%
Due to the absence of large-scale \{\emph{text}, \emph{3D asset}\}-paired datasets,
most methods exploit indirect supervision signals by rendering the current estimate of
the 3D asset into 2D images in multiple views and measuring 
similarity losses
between 
the rendered images and the given input text.
For measuring the similarity as losses,
%
the vision-language joint embedding space, \eg, CLIP~\cite{radford2021learning}, or 
text-conditional generative models, \eg, text-to-image diffusion model~\cite{rombach2021highresolution,saharia2022imagen}, are used.
This enables per-instance generation by optimization without any paired fully supervised data, \ie, 
synthesis-through-optimization.
%
%

With the synthesis-through-optimization framework, text-driven 3D asset generation methods are 
categorized into volume- and mesh-based 
methods.
Volume-based methods~\cite{poole2022dreamfusion,jain2021dreamfields,jiang2023avatarcraft,hong2022avatarclip,cao2023dreamavatar,huang2023dreamwaltz,wang2023prolificdreamer,metzer2022latent,wang2023scorejacobian,huang2023dreamtime} optimize the characteristics, \eg, occupancy, signed distance function (SDF), and color, of the points in a 3D space.
%
Mesh-based methods~\cite{youwang2022clipactor,oscar2022text2mesh,chen2023text2tex} model geometry with explicit 
meshes and generate vertex textures or texture maps. Using meshes allows rasterization for faster and more efficient rendering, in contrast to volumetric rendering used in the volume-based ones.
Also, meshes are well-compatible with graphics engines and suitable for texture transfer and animation.
This contrasts the volume representation that requires separate post-processing to extract mesh and unwrap a texture map by dedicated methods.
Thus, 3D designers prefer mesh representation due to its practicality.
%
Recently, hybrid methods
\cite{chen2023fantasia,lin2023magic3d}
were 
suggested, 
but they eventually perform re-meshing and texture unwrapping after the 3D volume optimization, which introduces substantial texture seams and loses editability. 
%

Our work chooses mesh representation
for synthesizing realistic or aesthetic 3D assets in high fidelity with practical compatibility. 
Specifically, we
focus on texture map synthesis, 
where we can obtain photorealistic renderings with fast and stable optimization. 

\paragraph{Text-driven Texture Map Synthesis}
While texture maps are the most commonly used for the graphics pipeline, there are only a few works that generate high-quality texture maps~\cite{siddiqui2022texturify,chen2023text2tex,chen2023fantasia,metzer2022latent,richardson2023texture}.
Text2Tex~\cite{chen2023text2tex} and TEXTure~\cite{richardson2023texture} are analogous, where they generate a RGB image using a pre-trained text and depth-conditioned diffusion model. 
%
Since they use color back-projection from the image onto the texture map, their texture maps are limited in 
diffuse RGB domain. Also, they need additional masking methods to carefully distinguish which part to update.
%
%
Latent-Paint~\cite{metzer2022latent} and TexFusion~\cite{cao2023texfusion} propose optimizing a latent feature texture map.
%
However, they can also decode RGB colors only due to the dependency of the pre-trained model that can only produce RGB, yielding limited photorealism.
Fantasia3D~\cite{chen2023fantasia} optimizes higher-dimensional physically-based rendering (PBR) materials and generates photorealistic text-driven 3D assets. 
They estimate per-point PBR material, rather than the spatially structured texture maps we use, and yield non-smooth 
textures.
Our $\ours$ optimizes the neural re-parameterized PBR material maps and obtains smooth and photorealistic 3D assets. 
Moreover, instead of generating an image and inpainting the texture map with low-dimensional colors, we directly synthesize the DC-PBR texture map;
%
thus, we do not perform re-meshing or texture unwrapping for each mesh, so it is naturally compatible with applications, \eg, texture transfer or mesh animation.

\section{\textbf{$\ours$}: Text-Driven PBR Texture Synthesis via Neural Re-parameterized Optimization}
\label{sec:method}

%
%

\begin{figure*}[t]
\centering
   \includegraphics[width=\linewidth]{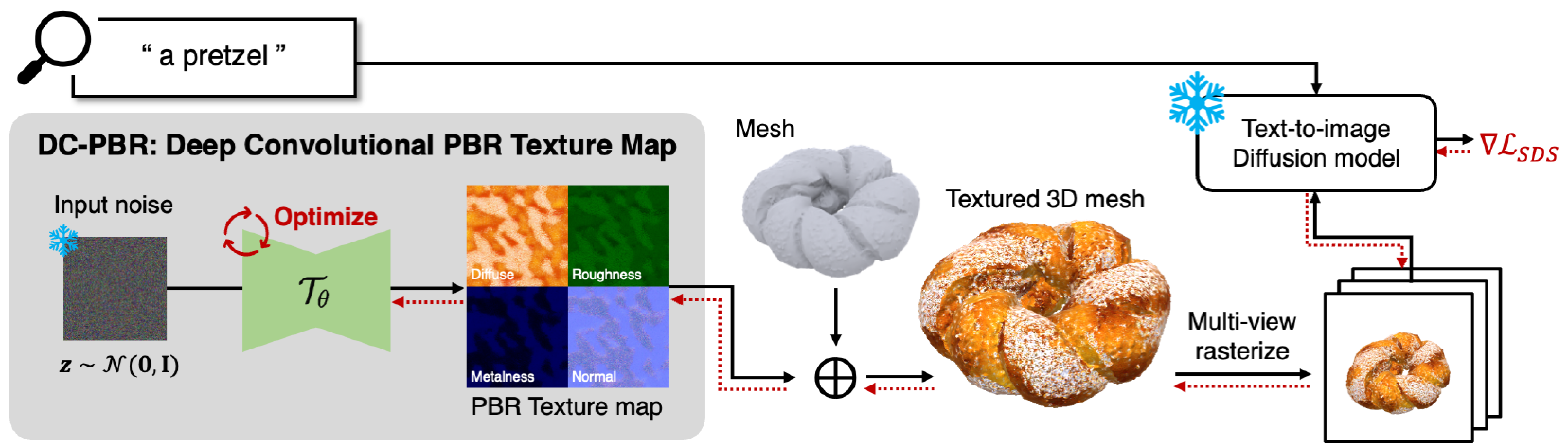}
   \caption{\textbf{$\ours$: Overall pipeline}.
   Given a 3D object mesh without a texture
   and a text describing the desired appearance of the mesh, $\ours$ synthesizes realistic PBR texture maps via synthesis-through-optimization. 
   We introduce DC-PBR, which parameterizes the PBR texture map into randomly initialized U-Net convolutional neural kernels.
   By performing texture mapping to texturize the given mesh, 
   we differentiably rasterize the textured mesh and obtain multi-view images, then compute the diffusion-guided loss. Note that $\ours$ optimizes the neural parameters of the U-Net rather than directly optimizing the pixel values of the texture map.
   %
   %
   %
   }   
\label{fig:pipeline} 
\end{figure*}

\subsection{Preliminary: Score-Distillation Sampling}\label{sec:preliminary}
The Score-Distillation Sampling (SDS)~\cite{poole2022dreamfusion} iteratively samples the 3D representation $\theta$ to generate an image that conforms to the input text description $y$.
%
Suppose there is a 3D representation, \eg, NeRF~\cite{mildenhall2020nerf}, parameterized as $\theta$, and we can render it into an image $\bx$ using a differentiable renderer, $g(\cdot)$, \ie, $\bx=g(\theta)$.
To perform SDS, we first perturb the rendered image $\bx=g(\theta)$ to make the noisy image $\bx_{t}$ by sampling a noise $\epsilon\sim\mathcal{N}(0, \bI)$ and a noising timestep $t\sim\mathcal{U}(0,1)$.
%
Initially, the rendered image $\bx$ would not look like an object described in the text prompt $y$.
Thus, given a pre-trained text-conditional noise estimator $\epsilon_{\phi}$, where $\phi$ denotes the parameters of the pre-trained diffusion model,
the error between 
%
the added noise $\epsilon$ and the text-conditioned estimated noise $\hat{\epsilon}_{\phi}(\bx_{t};y,t)$, \ie, $\lVert\hat{\epsilon}_{\phi}(\bx_{t};y,t)-\epsilon\rVert^2_2$, would be large.
On the contrary, if $\theta$ is well generated, and its rendering $\bx$ conforms to the text prompt and in the distribution of the training image, the error would be minimized.

Poole~\etal\cite{poole2022dreamfusion} formulate this intuition into an optimization problem, 
$\theta^{*}=\argmin_{\theta}L_\text{diff}(\phi,\bx{=}g(\theta))$,
where $L_\text{diff}(\phi,\bx{=}g(\theta))=\mathbb{E}_{t,\epsilon}[m(t)\lVert\hat{\epsilon}_{\phi}(\bx_{t};y,t)-\epsilon\rVert^2_2]$.
%
Thus, the update gradient for the 3D representation $\theta$ is written as:
\begin{equation}\label{eq:sds}
    \nabla_{\theta}\mathcal{L}_\text{SDS}(\phi,\bx)=\mathbb{E}_{t,\epsilon}\left[m(t)(\hat{\epsilon}_{\phi}(\bx_{t};y,t)-\epsilon)\frac{\partial\bx}{\partial\theta}\right]\,,
\end{equation}
where $m(t)$ denotes a weighting function conditioned on the diffusion noise timestep $t$.
This enables obtaining 3D from a text through 2D rendering even without any \{text, 3D\}-paired dataset.
We will use this 
gradient estimate to optimize the texture maps in a text-conditioned manner.

\subsection{Goal of \textbf{$\ours$}}
\label{sec:goal}
$\ours$ aims to synthesize high-fidelity physically-based rendering (PBR) texture maps for a given mesh and a text description so that the resulting texture maps visually conform to the text description.
Given a 3D mesh without texture, $\bM$, and
a text description $y$ describing the desired appearance of the mesh, our goal is to synthesize the PBR texture maps consisting of diffuse $\bK^\text{d}$, roughness \& metalness $\bK^\text{rm}$, and {detail surface} normal $\bK^\text{n}$ representations.
%
After synthesizing the PBR material texture maps, we can perform texture mapping to obtain a
%
text-conforming textured mesh. 
%

\subsection{DC-PBR: Deep Convolutional PBR Texture Map Re-parameterization}
\label{sec:reparam}

We propose the deep convolutional re-parameterization of PBR texture maps, $\dcpbr$,
$\mathcal{T}_\theta$.
Instead of the pixel value parameterization of texture maps, using $\dcpbr$ helps
{to sidestep the optimization difficulty of pixel-based representation, which will be discussed later in \Sref{sec:analysis}}.
%
We use a \emph{randomly initialized} convolutional U-Net with skip connections for $\mathcal{T}_\theta$ and use the randomly sampled code $\bz{\sim}\mathcal{N}(0,\bI)\in\mathbb{R}^{H\times W\times 3}$ as a fixed input, where
$H$ and $W$ are the height and width of the target texture maps, respectively, and $\bz$ is fixed during optimization.
%
With this, we re-parameterize the pixel-wise PBR parameters of the texture maps with the neural convolution kernels of the $\mathcal{T}_\theta$, \ie, 
$[\bK^{\text{d}}_{\theta}, \bK^{\text{rm}}_{\theta}, \bK^{\text{n}}_{\theta}]=\mathcal{T}_\theta(\bz)$, 
where
$\bK^{\text{d}}_{\theta},\bK^{\text{n}}_{\theta}\in\mathbb{R}^{H\times W\times 3}$,  $\bK^{\text{rm}}_{\theta}\in\mathbb{R}^{H\times W\times 2}$, and thus $\mathcal{T}_\theta(\bz)\in\mathbb{R}^{H\times W\times (3+2+3)}$.

\subsection{Text-driven 
DC-PBR
Optimization}
\label{sec:optimization}
%
Given a randomly initialized $\dcpbr$
$\mathcal{T}_\theta$ of the PBR texture maps, we perform an iterative optimization aid 
by the pre-trained text-to-image diffusion model.

\paragraph{Overall Pipeline}
We visualize the $\ours$ optimization pipeline in \Fref{fig:pipeline}.
At each iteration, we first feed 
the fixed noise $\bz$ to $\mathcal{T}_\theta$ and obtain the 
predictions of diffuse, roughness \& metalness
and normal maps; $\bK^{\text{d}}_{\theta}, \bK^{\text{rm}}_{\theta}$, and $\bK^{\text{n}}_{\theta}$.
We then rasterize the given mesh 
by texturing
with the obtained texture maps.
%
After rendering multi-view images of the textured mesh, 
we use the 
text-guided diffusion model to compute the update direction, $\nabla \mathcal{L}_\text{SDS}$, for the neural parameter $\theta$.

\paragraph{Rendering Mesh with PBR Texture Maps}
%
Given the output
PBR texture maps, \ie,
$\bK^{\text{d}}_{\theta}, \bK^{\text{rm}}_{\theta}, \bK^{\text{n}}_{\theta}$,
we texture the given mesh and perform differentiable rasterization to obtain rendered 
mesh images.
%
%
To render mesh surfaces, the diffuse $\bk^{\text{d}}_{\theta}\in\mathbb{R}^{3}$, 
roughness $\text{k}^{\text{r}}_{\theta}\in\mathbb{R}$,
metalness $\text{k}^{\text{m}}_{\theta}\in\mathbb{R}$,
and perturbing normal direction $\bk^{\text{n}}_{\theta}\in\mathbb{R}^{3}$ of a 3D surface point $\bp$ can be indexed from $\bK^{\text{d}}_{\theta}, \bK^{\text{rm}}_{\theta}$, and $\bK^{\text{n}}_{\theta}$,
using the \emph{uv} coordinates.
We can use the pre-defined \emph{uv} coordinates of the given mesh or perform unwrapping to generate the \emph{uv} coordinates.
%
We compute the specularity $\bk^{\text{s}}_{\theta}\in\mathbb{R}^{3}$ as: $\bk^{\text{s}}_{\theta}=0.04\cdot(1-\text{k}^{\text{m}}_{\theta})+\text{k}^{\text{m}}_{\theta}\cdot\bk^{\text{d}}_{\theta}$.
The rendered color $L$ of the mesh surface point $\bp$, seen from the view direction $\bomega$, can be computed using the rendering equation as:
\begin{equation}\label{eq:rendering}
L_\theta(\bp, \bomega) = \int_{\Omega} L_i(\bp, \bomega_i) f_\theta(\bp, \bomega_i, \bomega) \left(\bomega_i \cdot \bn_\theta\right) d\bomega_i,   
\end{equation}
where $\bomega_i$ denotes the incident light direction, $\Omega$ is a hemisphere around the perturbed surface normal $\bn_\theta$, and $L_i$ is the incident light from an off-the-shelf environment map.
{Also, $f_\theta(\bp, \bomega_i, \bomega)$ is the bidirectional reflectance distribution function (BRDF) of the material at 3D surface point $\bp$. 
We model the BRDF according to the PBR representation, $\bk^{\text{d}}_{\theta}$, $\bk^{\text{s}}_{\theta}$, and $\bk^{\text{n}}_{\theta}$, which is parameterized by our 
$\dcpbr$.
}

Using the renowned Cook-Torrance microfacet specular shading model~\cite{cook1982rendering}, we can decompose \Eref{eq:rendering} into the diffuse term $L_{\text{d}_{\theta}}(\bp)$ and the specular term $L_{\text{s}_{\theta}}(\bp, \bomega)$ as:
\begin{align}\label{eq:brdf}    
&L_\theta(\bp, \bomega) = L_{\text{d}_{\theta}}(\bp) + L_{\text{s}_{\theta}}(\bp, \bomega), \nonumber \\
&L_{\text{d}_{\theta}}(\bp) = \bk^{\text{d}}_{\theta}(1 - k^{\text{m}}_{\theta}) \int_{\Omega} L_i\left(\bp, \bomega_i\right)(\bomega_i \cdot \bn_\theta)d\bomega_i, \nonumber \\
&L_{\text{s}_{\theta}}\left(\bp, \bomega\right) = \int_{\Omega} \frac{D_\theta F_\theta G_\theta}{4(\bomega \cdot \bn_\theta)(\bomega_i \cdot \bn_\theta)} L_i(\bp, \bomega_i)(\bomega_i \cdot \bn_\theta)d\bomega_i, \nonumber 
\end{align}
where $D_\theta$, $F_\theta$, and $G_\theta$ denote the microfacet distribution, Fresnel term, and geometric attenuation function, respectively. 
Note that $D_\theta$, and $G_\theta$ are the functions of the generated $\text{k}^{\text{r}}_{\theta}$, and $F_\theta$ is the function of the specularity, $\bk^{\text{s}}_{\theta}$.
%

Iterating all the surface points, we obtain the image of the rendered mesh, $\bI_\theta$.
For simplicity, we denote the aforementioned rendering process for mesh $\bM$ as $\bI_\theta=\mathcal{R}^{\bM}(\bK^{\text{d}}_{\theta},\bK^{\text{rm}}_{\theta},\bK^{\text{n}}_{\theta})$,
where $\mathcal{R}^{\bM}(\cdot)$ denotes
the differentiable mesh rendering function, which we use
NVDiffRast~\cite{laine2020diffrast}.

\paragraph{Diffusion-guided 
DC-PBR
Optimization}
We obtain noisy PBR texture maps for the initial iteration of the optimization since the 
DC-PBR
$\mathcal{T}_\theta$ is randomly initialized.
%
We use the Score-Distillation Sampling (\Eref{eq:sds})
to iteratively update the neural re-parameterized PBR texture maps, $\mathcal{T}_\theta$.
Our optimization problem can be written as follows:
\begin{equation}
    \small
    \label{eq:objective}
    \theta^{*}{=}\argmin_{\theta} \mathbb{E}_{t,\epsilon}\left[\lVert\hat{\epsilon}_{\phi}(\mathcal{R}^{\bM}_{t}(\bK^{\text{d}}_{\theta},\bK^{\text{rm}}_{\theta},\bK^{\text{n}}_{\theta});y,t){-}\epsilon\rVert^2_2\right],
\end{equation}
where $\phi$ denotes the parameters of the pre-trained diffusion model, $\mathcal{R}^{\bM}_{t}(\bK^{\text{d}}_{\theta},\bK^{\text{rm}}_{\theta},\bK^{\text{n}}_{\theta})$ 
denotes the noisy image perturbed with forward diffusion process,
respectively.
We omit the $t$-dependent weighting function $m(t)$ for notation simplicity.
Given an image $\bI_\theta$ rendered from the textured mesh,
we compute the SDS update gradient
for updating the neural re-parameterized texture maps as follows:
\begin{gather*}
\label{eq:sds_ours}
    \nabla_{\theta}\mathcal{L}_\text{SDS}(\phi,\bI_\theta) =\mathbb{E}_{t,\epsilon}\left[(\hat{\epsilon}_{\phi}(\bI_{\theta,t};y,t)-\epsilon)\frac{\partial\bI_{\theta}}{\partial\theta}\right] \\
    = \mathbb{E}_{t,\epsilon}\left[\{\hat{\epsilon}_{\phi}(\mathcal{R}^{\bM}_{t}(\bK^{\text{d}}_{\theta},\bK^{\text{rm}}_{\theta},\bK^{\text{n}}_{\theta});y,t)-\epsilon\}\frac{\partial\bI_\theta}{\partial\theta}\right]\,.
\end{gather*}
%
%
%
The iterative update of DC-PBR $\mathcal{T}_\theta$ with $\nabla_{\theta}\mathcal{L}_\text{SDS}$ finally yields a solution
$\theta^{*}$, and we obtain high-quality PBR texture maps 
as: 
$[\bK^{\text{d}}_{\theta^{*}}, \bK^{\text{rm}}_{\theta^{*}},\bK^{\text{n}}_{\theta^{*}}]=\mathcal{T}_{{\theta}^{*}}(\bz)$.

\section{Analysis: Effect of 
the Deep Convolutional 
Re-parameterization for PBR Texture Maps}
\label{sec:analysis}

%
%

\subsection{Analysis of Fitting Behavior}
\label{sec:toy_example}

We observe that the SDS loss is noisy, including notable 
randomness.
%
To analyze, we first design a simple experiment focusing on the parameterization by excluding
the influence of the
randomness induced by the diffusion model.

\paragraph{Methods}
We compare the optimizations on pixel values and neural parameters as: 1)
\emph{Pixel Optimization}:
%
The most direct
way to fit an initial texture map $T\in{\mathbb{R}^{H \times W \times 3}}$ to the ground truth $\tilde{T}$
would be to optimize the pixel value of $T$ to minimize the error, \eg, L1 loss, as : ${T}^{*}=\argmin_{T}\lvert T-\tilde{T}\rvert$.
2) \emph{Neural Re-parameterized Optimization}:
%
Our method to fit a texture $T$ is to re-parameterize it with the neural parameters, \ie, $T{=}\mathcal{T}_\theta(\bz)$, where $\mathcal{T}_\theta(\cdot)$ is a randomly initialized convolutional U-Net with skip connections, and $\bz\sim\mathcal{N}(0,\bI)\in\mathbb{R}^{H\times W\times 3}$, which is fixed during the optimization.
Thus, the optimization problem is as follows:
${\theta}^{*}=\argmin_{\theta}\lvert \mathcal{T}_{\theta}(\bz)-\tilde{T}\rvert$.

\paragraph{Frequency Band Energy Analysis}
By comparing both methods, we see 
the characteristic differences of two representations: pixel parameters and deep convolutional re-parameterization.
%
In this analysis, we investigate the energies of the frequency components in the texture maps.
%
Given each iteration's texture map,
we conduct the spatial texture frequency (Fourier) analysis and compute
the energy components in five non-overlapping frequency bands from the lowest to highest frequencies.
See supplementary for details.

Figures~\ref{fig:sds_dip_freq}\colorref{a} and \colorref{4b} show the energy-iteration plot of both methods.
While the pixel value optimization fits all the frequency bands simultaneously
(\Fref{fig:sds_dip_freq}\colorref{a}),
the neural re-parameterized optimization 
fits the low-frequency components faster and defers to fit the high-frequency components later (\Fref{fig:sds_dip_freq}\colorref{b}), \ie, schedules the frequency.
%
Considering that lower-frequency bands mostly contain
the content of the image while 
highest-frequency bands mainly correlate with 
noises in the image,
%
%
we hypothesize that the scheduled frequency of neural re-parameterization helps the optimization focus more on the content of the texture map in the earlier iterations.
%
The texture map visualizations show the neural re-parameterized optimization fits the overall texture and skin tones, \ie, low-frequency, first in earlier iterations
and details later.
A similar observation in the natural image domain is reported in \cite{ulyanov2018dip,shi2022onmeasure}, and we further show that the consistent result also holds for the PBR representation.
%
On the other hand, the pixel 
optimization learns low-to-high frequencies simultaneously, 
which fits noise and texture signals jointly.
This yields undesirable optimization paths that may be harmful for sensitive losses like the SDS loss, which is further investigated as follows.

\begin{figure*}[t]
\centering
   \includegraphics[width=1.0\linewidth]{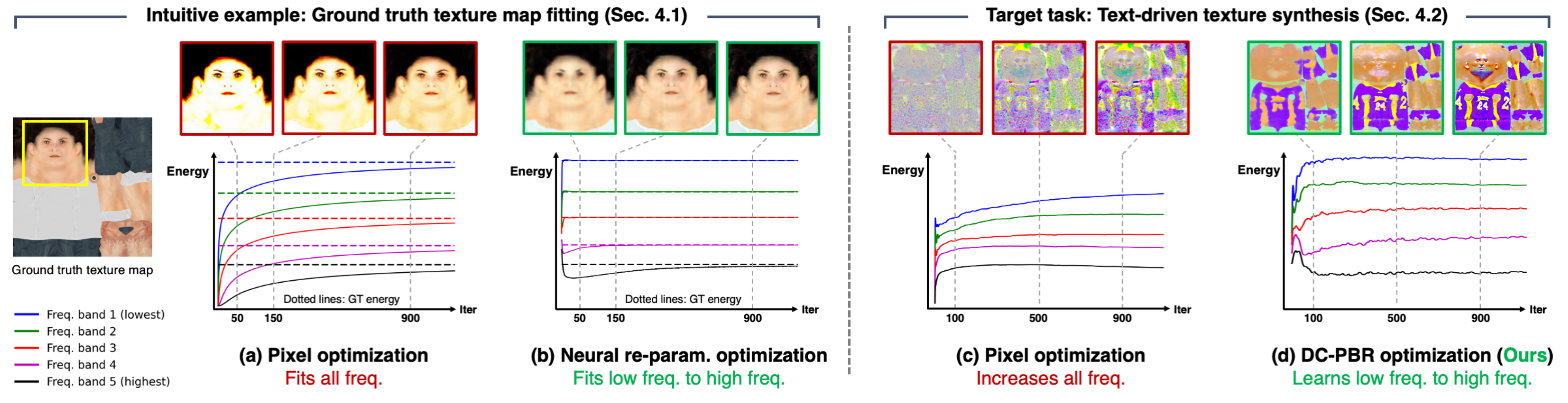}\vspace{-3mm}
   \caption{\textbf{Frequency scheduling of neural re-parameterized texture optimization}. 
   For each iteration, we investigate the energies of the frequency components of the reconstructed (a,b) / synthesized (c,d) texture maps.
   The pixel optimization (a,c) 
   fits and increases all frequency bands and suffers from fitting high-frequency texture contents from the initial stages, yielding degraded quality texture maps.
   In contrast,
   our proposed neural re-parameterization (b,d) naturally schedules which frequencies to focus on, thus obtaining coarse-to-fine texture synthesis
   with robustness to noisy supervision, \eg, SDS loss, and yielding high-quality texture maps.
   }\vspace{-3mm}   
\label{fig:sds_dip_freq} 
\end{figure*}

\subsection{Analysis of Optimization with the SDS Loss}
\label{sec:freq_schedule}

We investigate whether the observed frequency scheduling effect of our neural re-parameterization occurs in more complicated $\ours$ optimization with the SDS loss (\Eref{eq:objective}).
Note that the SDS loss is much noisier than the L1 loss in \Sref{sec:toy_example}.
The randomness in the sampled perturbation noise $\epsilon$, diffusion timestep $t$, and multi-view camera positions yield incoherent gradients in every optimization iteration.

Similar to the pixel 
optimization in \Sref{sec:toy_example}, 
we design the baseline pixel optimization for synthesizing PBR texture maps with the SDS loss as follows:\vspace{-1.5mm}
\begin{multline}
    \label{eq:baseline}[\bK^{\text{d}*},\bK^{\text{rm}*},\bK^{\text{n}*}] = 
    \argmin_{\bK^\text{d}, \bK^\text{rm},\bK^\text{n}}\\
    \mathbb{E}_{t,\epsilon}\left[\lVert\hat{\epsilon}_{\phi}(\mathcal{R}^{\bM}_{t}(\bK^{\text{d}},\bK^{\text{rm}},\bK^{\text{n}});y,t){-}\epsilon\rVert^2_2\right] + \mathcal{L}_\text{TV},
\end{multline}
where $\bK^\text{d},\bK^\text{rm},\bK^\text{n}$ denote the diffuse, roughness \& metalness, and normal maps, respectively. 
We also use the total variation 
$\mathcal{L}_\text{TV}$ for $\bK^\text{d}$ as a regularization 
to guide the smoothness of the local diffuse texture. This compensates for the lack of inductive bias in the pixel parameterization so that we can derive a stronger baseline to be compared.
%
%

Figures~\ref{fig:sds_dip_freq}\colorref{c} and \colorref{4d} compare the baseline pixel optimization (\Eref{eq:baseline}) and our proposed DC-PBR optimization (\Eref{eq:objective}) by plotting 
the frequency band energies of the PBR texture maps obtained in each iteration.
In \Fref{fig:sds_dip_freq}\colorref{c}, the baseline pixel optimization 
increases 
all frequency bands.
%
It fits noisy details from the SDS loss and yields significantly degraded texture maps.
On the contrary, in \Fref{fig:sds_dip_freq}\colorref{d}, our proposed neural re-parameterized optimization shows behaviors similar to 
those of \Fref{fig:sds_dip_freq}\colorref{b}. 
The neural re-parameterization of DC-PBR guides the optimization to learn
low-frequency bands faster than high-frequency noise, and later, mid-frequency bands follow.
%
As a result, interestingly, 
the texture maps are spontaneously synthesized in a coarse-to-fine manner
{perceptually}, 
where the overall structure and colors are learned first 
and the details, such as eyes and letters on the body, later.
%

Our neural re-parameterized optimization 
robustly filters out the high-frequency noise gradients from the SDS loss by its favorable frequency scheduling property.
{We postulate that this favorable property is 
induced by the architecture of the convolutional U-Net $\mathcal{T}_\theta$, consisting of a diverse composition of 
convolution kernels. 
The convolution kernel itself tends to 
learn favorable expressive local texture prior~\cite{gatys2015texture, ulyanov2018dip,soltanolkotabi2020denoising}, including smoothness.
Also, the stacked convolution mechanism that is repeatedly applied across the spatial domain with diverse compositions is analogous to other prior structures leveraging pattern recurrences of natural images, \eg, \cite{criminisi2003object, wang2018non, Barnes:2010:TGP, michaeli2013nonparametric}.}

\section{Experiments}
\label{sec:experiments}
%
%

\subsection{Qualitative Results}
\label{sec:qual}
\begin{figure*}[thbp]
    \centering
       \includegraphics[width=\linewidth]{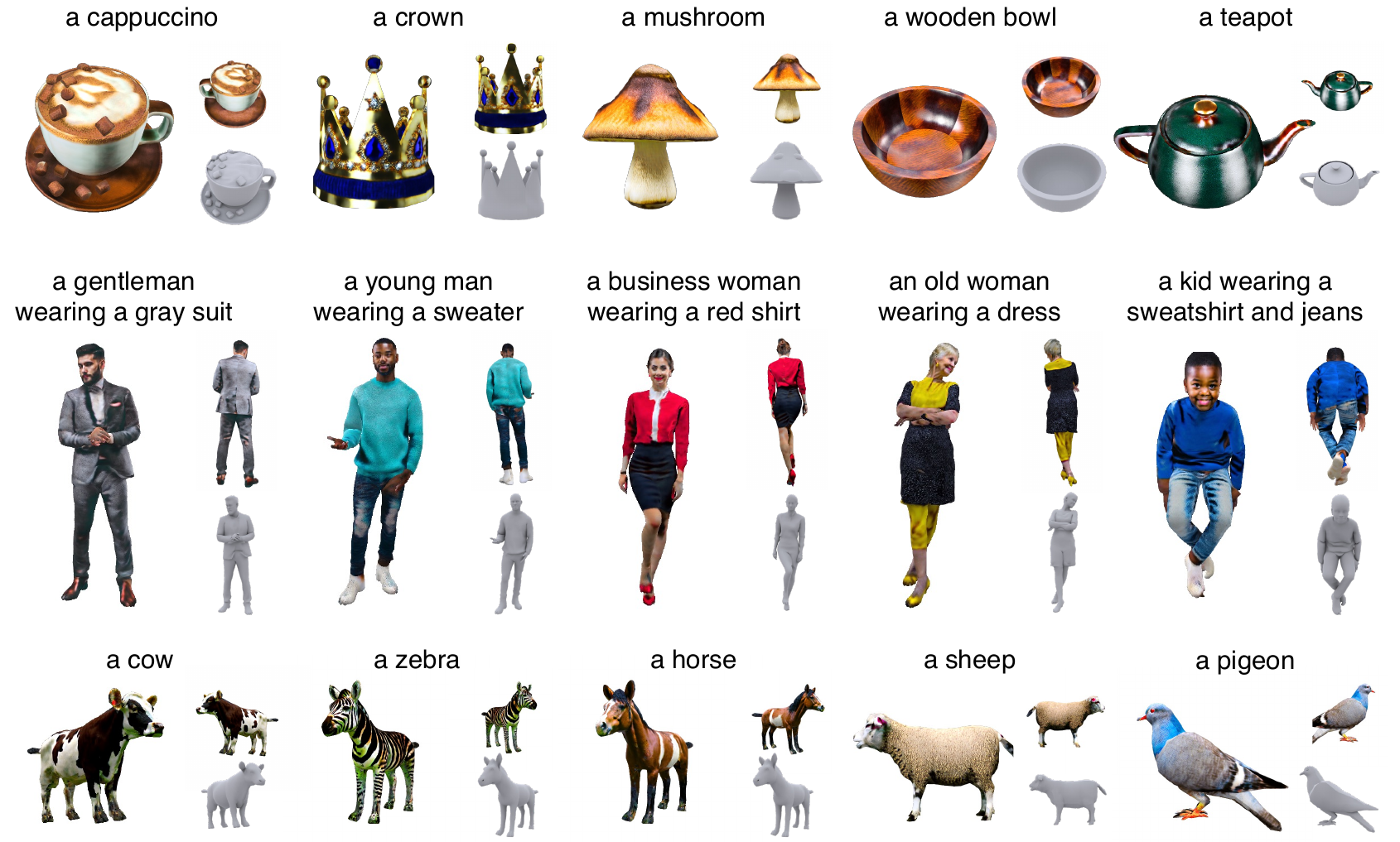}
       \caption{\textbf{Qualitative results}. 
       We take diverse 3D meshes from Objaverse~\cite{objaverse}, RenderPeople~\cite{renderpeople}, and SMAL~\cite{zuffi2017smal}, then synthesize texture maps with our manual text prompts.
       We visualize the original and rendered meshes with our synthesized PBR texture maps.
       %
       $\ours$ can model diverse material properties, \eg, the metallic surface of a crown, the rough surface of a mushroom, 
       realistic human skin tones, front-to-back appearance consistency, and complicated patterns of the animal's appearance.
       See supplementary material for more results.
       }   
    \label{fig:qual_all} 
    \vspace{3mm}
    \centering
    \includegraphics[width=\linewidth]{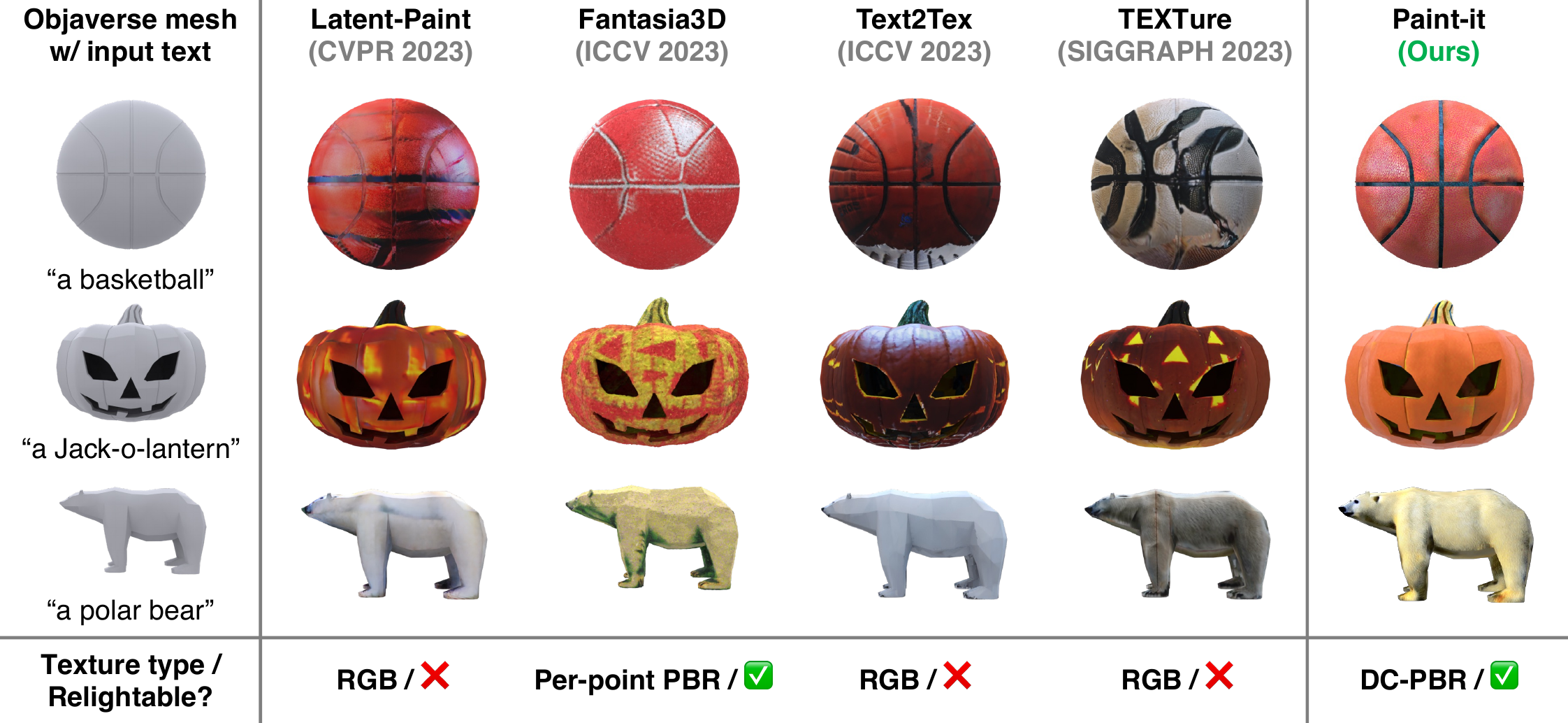}\vspace{-1.5mm}
    \caption{\textbf{Qualitative comparison}.
   We compare $\ours$ with recent competing methods~\cite{metzer2022latent,chen2023fantasia,chen2023text2tex,richardson2023texture}. 
   We script each method to synthesize textures for the subset of Objaverse~\cite{objaverse} meshes and compare the rendered quality of the textured meshes.
   Deep convolutional re-parameterization of the PBR texture maps helps $\ours$ synthesize a photorealistic and vivid appearance compared to other methods.
   }   
\label{fig:qual_comparison} 

\end{figure*}

%
%
In \Fref{fig:qual_all}, we visualize the rendered meshes
using $\ours$'s synthesized PBR texture maps for a given text prompt. 
To show the generalizability of the $\ours$ synthesis method,
we take the subsets of the large-scale mesh datasets: Objaverse~\cite{objaverse} and RenderPeople~\cite{renderpeople} for general objects and clothed humans.
For animals, we obtained the template meshes from the quadruped animal linear mesh model~\cite{zuffi2017smal}.
%
$\ours$ can generate photorealistic and vivid textures with material properties such as a mushroom's matte surface and a teapot's metallic surface.
%
By leveraging the strong generative prior from the pre-trained text-to-image diffusion model, $\ours$ faithfully 
distinguishes texture parts for skin and cloth.
Interestingly, $\ours$ can generate pseudo-stereoscopic effects, even though the given mesh surface is flat, \eg, the jewels and gems on a crown. 
We postulate this effect stems from our DC-PBR, where we synthesize the disentangled material properties
along with perturbing tangent space normals.
$\ours$ also supports the material control or texture transfer for the same input mesh and the relighting using different environment maps, thanks to the synthesized PBR texture maps 
(\Fref{fig:relight}).

\subsection{Comparison with Competing Methods}
\label{sec:competing}
\begin{figure}[t]
\centering
\includegraphics[width=0.9\linewidth]{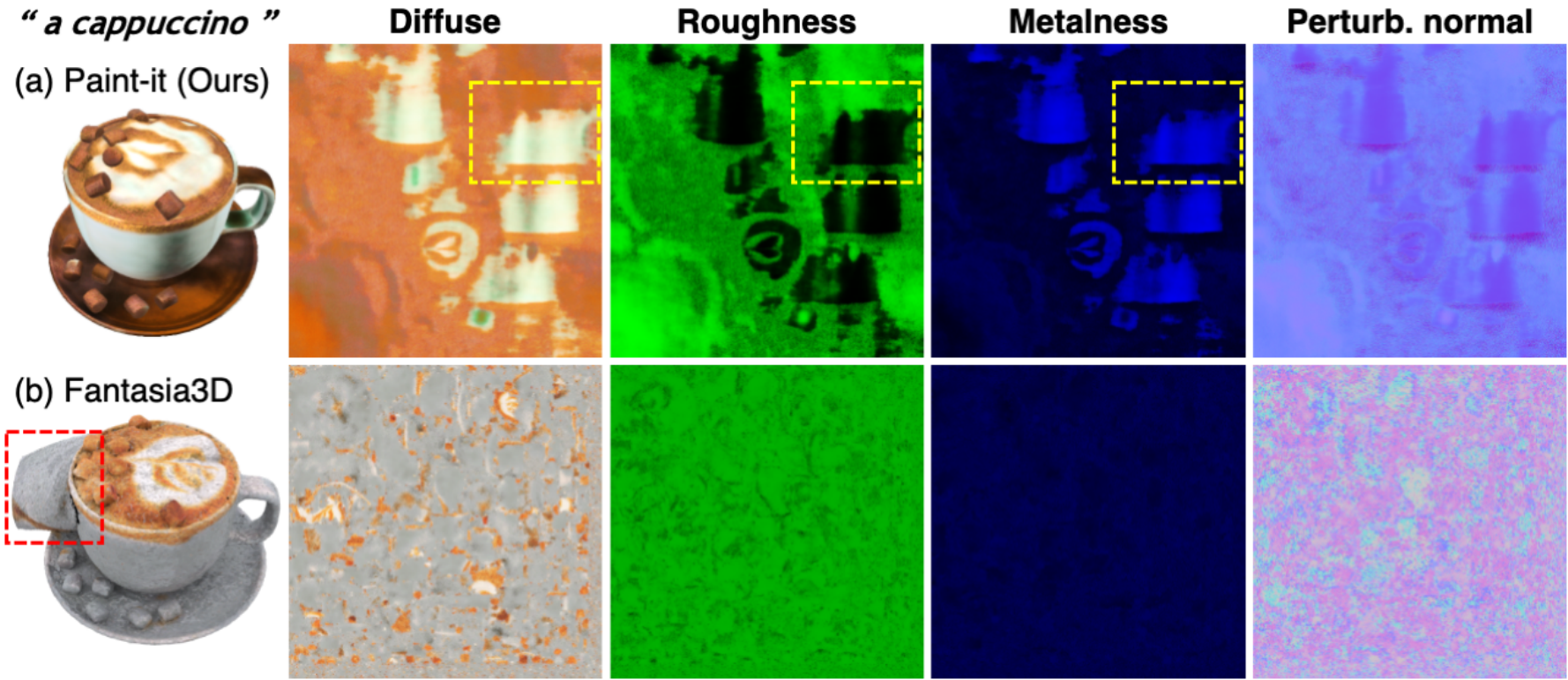}\vspace{-1.5mm}
   \caption{
   \textbf{PBR disentanglement results.} $\ours$ vs. Fantasia3D.
   }\vspace{-3mm}
\label{fig:reb_pbr}
\end{figure}

In \Fref{fig:qual_comparison}, we evaluate $\ours$ with recent text-driven mesh texture synthesis methods, 
Latent-Paint~\cite{metzer2022latent}, 
Fantasia3D~\cite{chen2023fantasia},
Text2Tex~\cite{chen2023text2tex}, and TEXTure~\cite{richardson2023texture}.
Texture maps are generated from each method on identical 3D meshes and text.
$\ours$ synthesizes more vivid, realistic, and consistent textures, 
compared to texture inpainting methods, Text2Tex and TEXTure. 
Specifically, they suffer from multi-view texture inconsistencies
and the baked lighting effects.
The back-projection of  
the generated RGB image 
onto the mesh and the limited diffuse texture representation could be the reason.
Latent-Paint synthesizes blurry texture and is also limited in diffuse texture. 
Fantasia3D learns a coordinate-based MLP to predict the per-point PBR materials, whereas we parameterize the full texture map globally. 
When backpropagating gradients, $\ours$ has a global effect over the full texture, while Fantasia3D is much more local. 
Given that SDS
is an ambiguous and noisy signal, the global gradient update of $\ours$ helps get a higher-quality, coherent appearance.
%
In \Fref{fig:reb_pbr}, $\ours$ obtain better-disentangled materials and specular properties,
while Fantasia3D fails to generate the mug's metallic (smooth) surface.
Also, Fantasia3D re-meshes the input mesh, destroying the geometry and obtaining implausible \emph{uv} mapping with substantial seams.

Following the protocol from Text2Tex~\cite{chen2023text2tex}, we report the Fr\'echet Inception Distance (FID)~\cite{heusel2017fid}.
Given untextured meshes from 
%
Objaverse~\cite{objaverse}, 
we scripted each method to synthesize texture maps from the
same text prompt.
Then, we render the textured
meshes in multi-views and compute the FID score. Please refer to Text2Tex for details. 
We also conduct a user study, requesting users
to rate the realism of 
samples synthesized with $\ours$ and competing methods. We got responses from 30 users.
%
\Tref{tab:quant} shows that Paint-it outperforms recent competing methods in terms of FID and user scores.
%
Only $\ours$ surpasses the score four (realistic), showing our superior realism and synthesis quality.

\begin{table}
    \centering
    \resizebox{0.9\linewidth}{!}{
    \begin{tabular}{lccccc}
        \toprule
             & Latent-Paint & Fantasia3D & Text2Tex & TEXTure & \textbf{Ours} \\
         \cmidrule{1-6}
            FID ($\downarrow$) & 41.11 & 58.79 & 37.89 & 38.40 & \textbf{34.46} \\
            User score ($\uparrow$) & 3.22 & 2.71 & 3.34 & 3.04 & \textbf{4.37} \\
        \bottomrule
    \end{tabular}
    }\vspace{-1.5mm}
    \caption{\textbf{Quantitative results on Objaverse subset.} 
    We evaluate the realism of the synthesized texture maps by measuring FID and user study.
    %
    $\ours$ outperforms the recent competing methods.
    }\vspace{-3mm}
    \label{tab:quant}
\end{table}

\subsection{Ablation studies}\label{sec:ablation}

\begin{figure}[t]
\centering
\includegraphics[width=\linewidth]{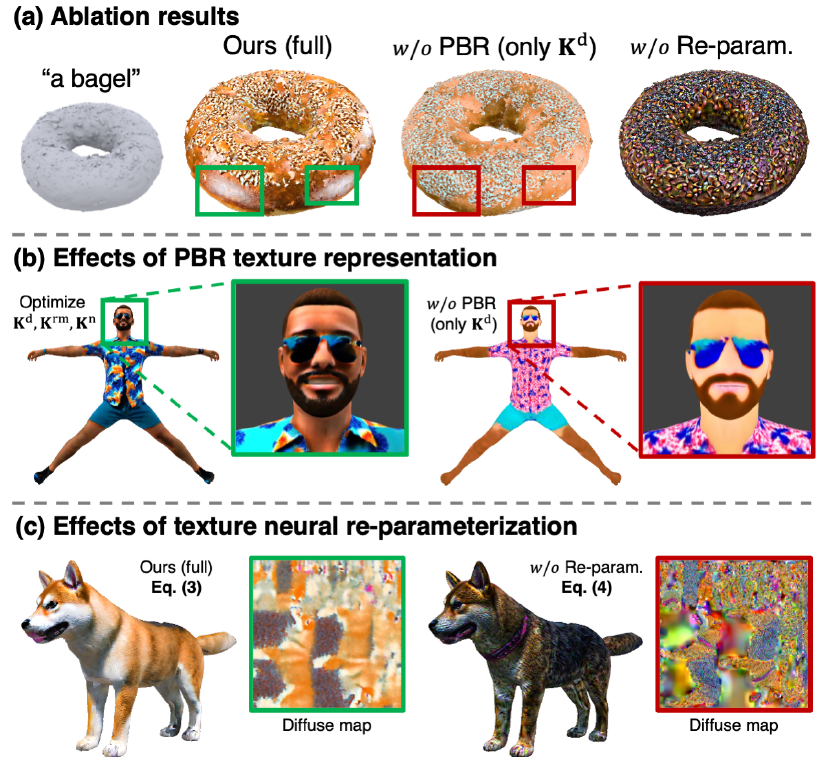}\vspace{-1.5mm}
   \caption{
   \textbf{Ablation study.}
   The proposed neural re-parameterization of PBR textures significantly enhances the visual qualities of the meshes, \eg, stereoscopic effects, realism, and texture consistency. 
   }\vspace{-4.5mm}
\label{fig:ablation}
\end{figure}

\paragraph{Effects of PBR Texture Representation}
First, we optimize only the diffuse texture map $\bK^\text{d}$ as other recent methods~\cite{chen2023text2tex,richardson2023texture}.
Simplifying the texture representation to model only the diffuse texture still generates a decent visual quality. However, compared to our full method, it is less realistic since it cannot model the reflection on the surface or stereoscopic effects.
%
We highlight the notable difference in the visual qualities of our full method and the diffuse-only optimization in \Fref{fig:ablation}\colorref{a}, \emph{w/o} PBR and in \Fref{fig:ablation}\colorref{b}.

\paragraph{Effects of Texture Neural Re-parameterization}
As discussed in \Sref{sec:analysis}, DC-PBR, \ie, neural re-parameterized optimization, naturally embodies the frequency scheduling for synthesizing textures.
%
While baseline optimization (\Eref{eq:baseline}) adopts a
regularization term
to avoid synthesizing noisy textures with high frequencies, 
it still introduces severely jittered textures (see \Fref{fig:ablation}\colorref{a}, \emph{w/o} Re-param., \& \Fref{fig:ablation}\colorref{c}).
%

\section{Discussion, Limitation, and Conclusion}
\label{sec:conclusion}

We present $\ours$, a text-based synthesis of physically-based rendering (PBR) texture maps for meshes. 
We propose the deep convolutional re-parameterization of 
PBR texture maps, 
which inherently 
eases and robustifies
the optimization
with the Score-Distillation Sampling. 
We show the performance and potential of the proposed method by synthesizing high-fidelity PBR texture maps for large-scale mesh datasets, including general objects, humans, and animals.

We expect $\ours$ can 
revolutionize 
the heuristic graphics pipelines, \eg, editing, relighting textures, and 
generating unlimited realistic 3D assets for production.
The current limitation of $\ours$ is the optimization time, which takes approximately 15$\sim$30 minutes per mesh. To further accelerate $\ours$, 
an efficient loss using the Consistency models~\cite{song2023consistency} would be helpful. 
Also, based on our synthesized texture maps 
for large-scale mesh datasets, 
curating a PBR texture map dataset and using it to train a feed-forward
generative model would be a promising future direction.

\scriptsize{
\paragraph{Acknowledgment}
We thank the members of AMILab~\cite{ami} and RVH group~\cite{rvh} for their helpful discussions and proofreading.
The project was made possible by funding from the Carl Zeiss Foundation.
This work is funded by the Deutsche Forschungsgemeinschaft (DFG, German
Research Foundation) - 409792180 (Emmy Noether Programme,
project: Real Virtual Humans), and the German Federal Ministry
of Education and Research (BMBF): T\"ubingen AI Center, FKZ: 01IS18039A.
Gerard Pons-Moll is a Professor at the University of T\"ubingen endowed by the Carl Zeiss Foundation, at the Department of Computer Science and a member of the Machine Learning Cluster of Excellence, EXC number 2064/1 – Project number 390727645.
Kim Youwang and Tae-Hyun Oh were supported by Institute of Information \& communications Technology Planning \& Evaluation (IITP) grant funded by the Korea government(MSIT) (No.RS-2023-00225630, Development of Artificial Intelligence for Text-based 3D Movie Generation; 
No.2022-0-00290, Visual Intelligence for Space-Time Understanding and Generation based on Multi-layered Visual Common Sense;
No.2021-0-02068, Artificial Intelligence Innovation Hub).}

{
    \small
    \bibliographystyle{ieeenat_fullname}
    \bibliography{main}

\begin{thebibliography}{65}
\providecommand{\natexlab}[1]{#1}
\providecommand{\url}[1]{\texttt{#1}}
\expandafter\ifx\csname urlstyle\endcsname\relax
  \providecommand{\doi}[1]{doi: #1}\else
  \providecommand{\doi}{doi: \begingroup \urlstyle{rm}\Url}\fi

\bibitem[ami()]{ami}
\href{https://ami.postech.ac.kr/members}{https://ami.postech.ac.kr/members}.

\bibitem[rvh()]{rvh}
\href{http://virtualhumans.mpi-inf.mpg.de/people.html}{http://virtualhumans.mpi-inf.mpg.de/people.html}.

\bibitem[ren(2023)]{renderpeople}
\href{https://renderpeople.com/}{https://renderpeople.com/}, 2023.

\bibitem[Barnes et~al.(2010)Barnes, Shechtman, Goldman, and Finkelstein]{Barnes:2010:TGP}
Connelly Barnes, Eli Shechtman, Dan~B Goldman, and Adam Finkelstein.
\newblock The generalized {PatchMatch} correspondence algorithm.
\newblock In \emph{European Conference on Computer Vision (ECCV)}, 2010.

\bibitem[Biggs et~al.(2020)Biggs, Boyne, Charles, Fitzgibbon, and Cipolla]{biggs2020wldo}
Benjamin Biggs, Oliver Boyne, James Charles, Andrew Fitzgibbon, and Roberto Cipolla.
\newblock {W}ho left the dogs out?: {3D} animal reconstruction with expectation maximization in the loop.
\newblock In \emph{European Conference on Computer Vision (ECCV)}, 2020.

\bibitem[Cao et~al.(2023{\natexlab{a}})Cao, Kreis, Fidler, Sharp, and Yin]{cao2023texfusion}
Tianshi Cao, Karsten Kreis, Sanja Fidler, Nicholas Sharp, and KangXue Yin.
\newblock Texfusion: Synthesizing 3d textures with text-guided image diffusion models.
\newblock In \emph{IEEE International Conference on Computer Vision (ICCV)}, 2023{\natexlab{a}}.

\bibitem[Cao et~al.(2023{\natexlab{b}})Cao, Cao, Han, Shan, and Wong]{cao2023dreamavatar}
Yukang Cao, Yan-Pei Cao, Kai Han, Ying Shan, and Kwan-Yee~K. Wong.
\newblock Dreamavatar: Text-and-shape guided 3d human avatar generation via diffusion models.
\newblock \emph{arXiv preprint, arxiv:2304.00916}, 2023{\natexlab{b}}.

\bibitem[Chan et~al.(2022)Chan, Lin, Chan, Nagano, Pan, Mello, Gallo, Guibas, Tremblay, Khamis, Karras, and Wetzstein]{chan2022eg3d}
Eric~R. Chan, Connor~Z. Lin, Matthew~A. Chan, Koki Nagano, Boxiao Pan, Shalini~De Mello, Orazio Gallo, Leonidas Guibas, Jonathan Tremblay, Sameh Khamis, Tero Karras, and Gordon Wetzstein.
\newblock Efficient geometry-aware {3D} generative adversarial networks.
\newblock In \emph{IEEE Conference on Computer Vision and Pattern Recognition (CVPR)}, 2022.

\bibitem[Chen et~al.(2023{\natexlab{a}})Chen, Siddiqui, Lee, Tulyakov, and Nie{\ss}ner]{chen2023text2tex}
Dave~Zhenyu Chen, Yawar Siddiqui, Hsin-Ying Lee, Sergey Tulyakov, and Matthias Nie{\ss}ner.
\newblock Text2tex: Text-driven texture synthesis via diffusion models.
\newblock In \emph{IEEE International Conference on Computer Vision (ICCV)}, 2023{\natexlab{a}}.

\bibitem[Chen et~al.(2023{\natexlab{b}})Chen, Chen, Jiao, and Jia]{chen2023fantasia}
Rui Chen, Yongwei Chen, Ningxin Jiao, and Kui Jia.
\newblock Fantasia3d: Disentangling geometry and appearance for high-quality text-to-3d content creation.
\newblock In \emph{IEEE International Conference on Computer Vision (ICCV)}, 2023{\natexlab{b}}.

\bibitem[Chen et~al.(2022{\natexlab{a}})Chen, Jiang, Song, Yang, Black, Geiger, and Hilliges]{chen2022gdna}
Xu Chen, Tianjian Jiang, Jie Song, Jinlong Yang, Michael Black, Andreas Geiger, and Otmar Hilliges.
\newblock gdna: Towards generative detailed neural avatars.
\newblock In \emph{IEEE Conference on Computer Vision and Pattern Recognition (CVPR)}, 2022{\natexlab{a}}.

\bibitem[Chen et~al.(2022{\natexlab{b}})Chen, Chen, Lei, Zhang, and Jia]{chen2022tango}
Yongwei Chen, Rui Chen, Jiabao Lei, Yabin Zhang, and Kui Jia.
\newblock Tango: Text-driven photorealistic and robust 3d stylization via lighting decomposition.
\newblock In \emph{Advances in Neural Information Processing Systems (NeurIPS)}, 2022{\natexlab{b}}.

\bibitem[Cho et~al.(2022)Cho, Youwang, and Oh]{cho2022FastMETRO}
Junhyeong Cho, Kim Youwang, and Tae-Hyun Oh.
\newblock Cross-attention of disentangled modalities for 3d human mesh recovery with transformers.
\newblock In \emph{European Conference on Computer Vision (ECCV)}, 2022.

\bibitem[Cook and Torrance(1982)]{cook1982rendering}
R.~L. Cook and K.~E. Torrance.
\newblock A reflectance model for computer graphics.
\newblock \emph{ACM Transactions on Graphics (SIGGRAPH)}, 1\penalty0 (1), 1982.

\bibitem[Criminisi et~al.(2003)Criminisi, Perez, and Toyama]{criminisi2003object}
Antonio Criminisi, Patrick Perez, and Kentaro Toyama.
\newblock Object removal by exemplar-based inpainting.
\newblock In \emph{IEEE Conference on Computer Vision and Pattern Recognition (CVPR)}, 2003.

\bibitem[Deitke et~al.(2023)Deitke, Schwenk, Salvador, Weihs, Michel, VanderBilt, Schmidt, Ehsani, Kembhavi, and Farhadi]{objaverse}
Matt Deitke, Dustin Schwenk, Jordi Salvador, Luca Weihs, Oscar Michel, Eli VanderBilt, Ludwig Schmidt, Kiana Ehsani, Aniruddha Kembhavi, and Ali Farhadi.
\newblock Objaverse: A universe of annotated 3d objects.
\newblock In \emph{IEEE Conference on Computer Vision and Pattern Recognition (CVPR)}, 2023.

\bibitem[Dong et~al.(2023)Dong, Chen, Jinlong~Yang, Hilliges, and Geiger]{dong2023ag3d}
Zijian Dong, Xu Chen, Michael~J.Black Jinlong~Yang, Otmar Hilliges, and Andreas Geiger.
\newblock {AG3D}: Learning to generate {3D} avatars from {2D} image collections.
\newblock In \emph{IEEE International Conference on Computer Vision (ICCV)}, 2023.

\bibitem[Feng et~al.(2021)Feng, Feng, Black, and Bolkart]{feng2021deca}
Yao Feng, Haiwen Feng, Michael~J. Black, and Timo Bolkart.
\newblock Learning an animatable detailed {3D} face model from in-the-wild images.
\newblock \emph{ACM Transactions on Graphics (SIGGRAPH)}, 40\penalty0 (8), 2021.

\bibitem[Gatys et~al.(2015)Gatys, Ecker, and Bethge]{gatys2015texture}
Leon Gatys, Alexander~S Ecker, and Matthias Bethge.
\newblock Texture synthesis using convolutional neural networks.
\newblock In \emph{Advances in Neural Information Processing Systems (NeurIPS)}, 2015.

\bibitem[Goel et~al.(2023)Goel, Pavlakos, Rajasegaran, Kanazawa*, and Malik*]{goel2023humans}
Shubham Goel, Georgios Pavlakos, Jathushan Rajasegaran, Angjoo Kanazawa*, and Jitendra Malik*.
\newblock Humans in 4{D}: Reconstructing and tracking humans with transformers.
\newblock In \emph{IEEE International Conference on Computer Vision (ICCV)}, 2023.

\bibitem[Guzov et~al.(2021)Guzov, Mir, Sattler, and Pons-Moll]{vladimir2021hps}
Vladimir Guzov, Aymen Mir, Torsten Sattler, and Gerard Pons-Moll.
\newblock Human poseitioning system (hps): 3d human pose estimation and self-localization in large scenes from body-mounted sensors.
\newblock In \emph{IEEE Conference on Computer Vision and Pattern Recognition (CVPR)}, 2021.

\bibitem[Heckel and Soltanolkotabi(2020)]{soltanolkotabi2020denoising}
Reinhard Heckel and Mahdi Soltanolkotabi.
\newblock Denoising and regularization via exploiting the structural bias of convolutional generators.
\newblock In \emph{International Conference on Learning Representations (ICLR)}, 2020.

\bibitem[Heusel et~al.(2017)Heusel, Ramsauer, Unterthiner, Nessler, and Hochreiter]{heusel2017fid}
Martin Heusel, Hubert Ramsauer, Thomas Unterthiner, Bernhard Nessler, and Sepp Hochreiter.
\newblock Gans trained by a two time-scale update rule converge to a local nash equilibrium.
\newblock In \emph{Advances in Neural Information Processing Systems (NeurIPS)}, 2017.

\bibitem[Hong et~al.(2022)Hong, Zhang, Pan, Cai, Yang, and Liu]{hong2022avatarclip}
Fangzhou Hong, Mingyuan Zhang, Liang Pan, Zhongang Cai, Lei Yang, and Ziwei Liu.
\newblock Avatarclip: Zero-shot text-driven generation and animation of 3d avatars.
\newblock \emph{ACM Transactions on Graphics (SIGGRAPH)}, 41\penalty0 (4):\penalty0 1--19, 2022.

\bibitem[Huang et~al.(2023{\natexlab{a}})Huang, Wang, Shi, Qi, Zha, and Zhang]{huang2023dreamtime}
Yukun Huang, Jianan Wang, Yukai Shi, Xianbiao Qi, Zheng-Jun Zha, and Lei Zhang.
\newblock Dreamtime: An improved optimization strategy for text-to-3d content creation.
\newblock \emph{arXiv preprint, arxiv:2306.12422}, 2023{\natexlab{a}}.

\bibitem[Huang et~al.(2023{\natexlab{b}})Huang, Wang, Zeng, Cao, Qi, Shi, Zha, and Zhang]{huang2023dreamwaltz}
Yukun Huang, Jianan Wang, Ailing Zeng, He Cao, Xianbiao Qi, Yukai Shi, Zheng-Jun Zha, and Lei Zhang.
\newblock Dreamwaltz: Make a scene with complex 3d animatable avatars.
\newblock \emph{arXiv preprint, arxiv:2305.12529}, 2023{\natexlab{b}}.

\bibitem[Jain et~al.(2022)Jain, Mildenhall, Barron, Abbeel, and Poole]{jain2021dreamfields}
Ajay Jain, Ben Mildenhall, Jonathan~T. Barron, Pieter Abbeel, and Ben Poole.
\newblock Zero-shot text-guided object generation with dream fields.
\newblock In \emph{IEEE Conference on Computer Vision and Pattern Recognition (CVPR)}, 2022.

\bibitem[Jiang et~al.(2023)Jiang, Wang, Zhang, Chai, He, Chen, and Liao]{jiang2023avatarcraft}
Ruixiang Jiang, Can Wang, Jingbo Zhang, Menglei Chai, Mingming He, Dongdong Chen, and Jing Liao.
\newblock Avatarcraft: Transforming text into neural human avatars with parameterized shape and pose control.
\newblock In \emph{IEEE International Conference on Computer Vision (ICCV)}, 2023.

\bibitem[Kim et~al.(2024)Kim, Youwang, and Oh]{kim2024fprf}
GeonU Kim, Kim Youwang, and Tae-Hyun Oh.
\newblock {FPRF}: Feed-forward photorealistic style transfer of large-scale {3D} neural radiance fields.
\newblock In \emph{AAAI Conference on Artificial Intelligence (AAAI)}, 2024.

\bibitem[Kim et~al.(2022)Kim, Kim, and Choi]{kim2022flame}
Jihoon Kim, Jiseob Kim, and Sungjoon Choi.
\newblock Flame: Free-form language-based motion synthesis \& editing.
\newblock In \emph{AAAI Conference on Artificial Intelligence (AAAI)}, 2022.

\bibitem[Laine et~al.(2020)Laine, Hellsten, Karras, Seol, Lehtinen, and Aila]{laine2020diffrast}
Samuli Laine, Janne Hellsten, Tero Karras, Yeongho Seol, Jaakko Lehtinen, and Timo Aila.
\newblock Modular primitives for high-performance differentiable rendering.
\newblock \emph{ACM Transactions on Graphics (SIGGRAPH)}, 39\penalty0 (6), 2020.

\bibitem[Li et~al.(2023)Li, M\"uller, Evans, Taylor, Unberath, Liu, and Lin]{li2023neuralangelo}
Zhaoshuo Li, Thomas M\"uller, Alex Evans, Russell~H Taylor, Mathias Unberath, Ming-Yu Liu, and Chen-Hsuan Lin.
\newblock Neuralangelo: High-fidelity neural surface reconstruction.
\newblock In \emph{IEEE Conference on Computer Vision and Pattern Recognition (CVPR)}, 2023.

\bibitem[Lin et~al.(2023)Lin, Gao, Tang, Takikawa, Zeng, Huang, Kreis, Fidler, Liu, and Lin]{lin2023magic3d}
Chen-Hsuan Lin, Jun Gao, Luming Tang, Towaki Takikawa, Xiaohui Zeng, Xun Huang, Karsten Kreis, Sanja Fidler, Ming-Yu Liu, and Tsung-Yi Lin.
\newblock Magic3d: High-resolution text-to-3d content creation.
\newblock In \emph{IEEE Conference on Computer Vision and Pattern Recognition (CVPR)}, 2023.

\bibitem[Ma et~al.(2020)Ma, Yang, Ranjan, Pujades, Pons-Moll, Tang, and Black]{ma20autoenclother}
Qianli Ma, Jinlong Yang, Anurag Ranjan, Sergi Pujades, Gerard Pons-Moll, Siyu Tang, and Michael Black.
\newblock Learning to dress 3d people in generative clothing.
\newblock In \emph{IEEE Conference on Computer Vision and Pattern Recognition (CVPR)}, 2020.

\bibitem[Metzer et~al.(2023)Metzer, Richardson, Patashnik, Giryes, and Cohen-Or]{metzer2022latent}
Gal Metzer, Elad Richardson, Or Patashnik, Raja Giryes, and Daniel Cohen-Or.
\newblock Latent-nerf for shape-guided generation of 3d shapes and textures.
\newblock In \emph{IEEE Conference on Computer Vision and Pattern Recognition (CVPR)}, 2023.

\bibitem[Michaeli and Irani(2013)]{michaeli2013nonparametric}
Tomer Michaeli and Michal Irani.
\newblock Nonparametric blind super-resolution.
\newblock In \emph{IEEE International Conference on Computer Vision (ICCV)}, 2013.

\bibitem[Michel et~al.(2022)Michel, Bar-On, Liu, Benaim, and Hanocka]{oscar2022text2mesh}
Oscar Michel, Roi Bar-On, Richard Liu, Sagie Benaim, and Rana Hanocka.
\newblock Text2mesh: Text-driven neural stylization for meshes.
\newblock In \emph{IEEE Conference on Computer Vision and Pattern Recognition (CVPR)}, 2022.

\bibitem[Mildenhall et~al.(2020)Mildenhall, Srinivasan, Tancik, Barron, Ramamoorthi, and Ng]{mildenhall2020nerf}
Ben Mildenhall, Pratul~P. Srinivasan, Matthew Tancik, Jonathan~T. Barron, Ravi Ramamoorthi, and Ren Ng.
\newblock Nerf: Representing scenes as neural radiance fields for view synthesis.
\newblock In \emph{European Conference on Computer Vision (ECCV)}, 2020.

\bibitem[M\"uller et~al.(2022)M\"uller, Evans, Schied, and Keller]{mueller2022ingp}
Thomas M\"uller, Alex Evans, Christoph Schied, and Alexander Keller.
\newblock Instant neural graphics primitives with a multiresolution hash encoding.
\newblock \emph{ACM Transactions on Graphics (SIGGRAPH)}, 41\penalty0 (4):\penalty0 102:1--102:15, 2022.

\bibitem[Park et~al.(2019)Park, Florence, Straub, Newcombe, and Lovegrove]{park2019deepsdf}
Jeong~Joon Park, Peter Florence, Julian Straub, Richard Newcombe, and Steven Lovegrove.
\newblock Deepsdf: Learning continuous signed distance functions for shape representation.
\newblock In \emph{IEEE Conference on Computer Vision and Pattern Recognition (CVPR)}, 2019.

\bibitem[Pavlakos et~al.(2019)Pavlakos, Choutas, Ghorbani, Bolkart, Osman, Tzionas, and Black]{pavlakos2019smplx}
Georgios Pavlakos, Vasileios Choutas, Nima Ghorbani, Timo Bolkart, Ahmed A.~A. Osman, Dimitrios Tzionas, and Michael~J. Black.
\newblock Expressive body capture: {3D} hands, face, and body from a single image.
\newblock In \emph{IEEE Conference on Computer Vision and Pattern Recognition (CVPR)}, 2019.

\bibitem[Poole et~al.(2022)Poole, Jain, Barron, and Mildenhall]{poole2022dreamfusion}
Ben Poole, Ajay Jain, Jonathan~T. Barron, and Ben Mildenhall.
\newblock Dreamfusion: Text-to-3d using 2d diffusion.
\newblock In \emph{International Conference on Learning Representations (ICLR)}, 2022.

\bibitem[Radford et~al.(2021)Radford, Kim, Hallacy, Ramesh, Goh, Agarwal, Sastry, Askell, Mishkin, Clark, Krueger, and Sutskever]{radford2021learning}
Alec Radford, Jong~Wook Kim, Chris Hallacy, Aditya Ramesh, Gabriel Goh, Sandhini Agarwal, Girish Sastry, Amanda Askell, Pamela Mishkin, Jack Clark, Gretchen Krueger, and Ilya Sutskever.
\newblock Learning transferable visual models from natural language supervision.
\newblock In \emph{International Conference on Machine Learning (ICML)}, 2021.

\bibitem[Rempe et~al.(2021)Rempe, Birdal, Hertzmann, Yang, Sridhar, and Guibas]{rempe2021humor}
Davis Rempe, Tolga Birdal, Aaron Hertzmann, Jimei Yang, Srinath Sridhar, and Leonidas~J. Guibas.
\newblock Humor: 3d human motion model for robust pose estimation.
\newblock In \emph{IEEE International Conference on Computer Vision (ICCV)}, 2021.

\bibitem[Richardson et~al.(2023)Richardson, Metzer, Alaluf, Giryes, and Cohen-Or]{richardson2023texture}
Elad Richardson, Gal Metzer, Yuval Alaluf, Raja Giryes, and Daniel Cohen-Or.
\newblock Texture: Text-guided texturing of 3d shapes.
\newblock \emph{ACM Transactions on Graphics (SIGGRAPH)}, 2023.

\bibitem[Rombach et~al.(2022)Rombach, Blattmann, Lorenz, Esser, and Ommer]{rombach2021highresolution}
Robin Rombach, Andreas Blattmann, Dominik Lorenz, Patrick Esser, and Björn Ommer.
\newblock High-resolution image synthesis with latent diffusion models.
\newblock In \emph{IEEE Conference on Computer Vision and Pattern Recognition (CVPR)}, 2022.

\bibitem[Saharia et~al.(2022)Saharia, Chan, Saxena, Li, Whang, Denton, Ghasemipour, Ayan, Mahdavi, Lopes, Salimans, Ho, Fleet, and Norouzi]{saharia2022imagen}
Chitwan Saharia, William Chan, Saurabh Saxena, Lala Li, Jay Whang, Emily~L. Denton, Seyed Kamyar~Seyed Ghasemipour, Burcu~Karagol Ayan, Seyedeh~Sara Mahdavi, Raphael~Gontijo Lopes, Tim Salimans, Jonathan Ho, David Fleet, and Mohammad Norouzi.
\newblock Photorealistic text-to-image diffusion models with deep language understanding.
\newblock In \emph{Advances in Neural Information Processing Systems (NeurIPS)}, 2022.

\bibitem[Shen et~al.(2021)Shen, Gao, Yin, Liu, and Fidler]{shen2021dmtet}
Tianchang Shen, Jun Gao, Kangxue Yin, Ming-Yu Liu, and Sanja Fidler.
\newblock Deep marching tetrahedra: a hybrid representation for high-resolution 3d shape synthesis.
\newblock In \emph{Advances in Neural Information Processing Systems (NeurIPS)}, 2021.

\bibitem[Shi et~al.(2022)Shi, Mettes, Maji, and Snoek]{shi2022onmeasure}
Zenglin Shi, Pascal Mettes, Subhransu Maji, and Cees G~M Snoek.
\newblock On measuring and controlling the spectral bias of the deep image prior.
\newblock \emph{International Journal of Computer Vision}, 2022.

\bibitem[Siddiqui et~al.(2022)Siddiqui, Thies, Ma, Shan, Nie{\ss}ner, and Dai]{siddiqui2022texturify}
Yawar Siddiqui, Justus Thies, Fangchang Ma, Qi Shan, Matthias Nie{\ss}ner, and Angela Dai.
\newblock Texturify: Generating textures on 3d shape surfaces.
\newblock In \emph{European Conference on Computer Vision (ECCV)}, 2022.

\bibitem[Song et~al.(2023)Song, Dhariwal, Chen, and Sutskever]{song2023consistency}
Yang Song, Prafulla Dhariwal, Mark Chen, and Ilya Sutskever.
\newblock Consistency models.
\newblock In \emph{International Conference on Machine Learning (ICML)}, 2023.

\bibitem[Sung-Bin et~al.(2024)Sung-Bin, Hyun, Hong, Nam, Ju, and Oh]{sungbin2024laughtalk}
Kim Sung-Bin, Lee Hyun, Da~Hye Hong, Suekyeong Nam, Janghoon Ju, and Tae-Hyun Oh.
\newblock Laughtalk: Expressive 3d talking head generation with laughter, 2024.

\bibitem[Tevet et~al.(2023)Tevet, Raab, Gordon, Shafir, Cohen-or, and Bermano]{tevet2023human}
Guy Tevet, Sigal Raab, Brian Gordon, Yoni Shafir, Daniel Cohen-or, and Amit~Haim Bermano.
\newblock Human motion diffusion model.
\newblock In \emph{International Conference on Learning Representations (ICLR)}, 2023.

\bibitem[Ulyanov et~al.(2018)Ulyanov, Vedaldi, and Lempitsky]{ulyanov2018dip}
Dmitry Ulyanov, Andrea Vedaldi, and Victor Lempitsky.
\newblock Deep image prior.
\newblock In \emph{IEEE Conference on Computer Vision and Pattern Recognition (CVPR)}, 2018.

\bibitem[Wang et~al.(2023{\natexlab{a}})Wang, Du, Li, Yeh, and Shakhnarovich]{wang2023scorejacobian}
Haochen Wang, Xiaodan Du, Jiahao Li, Raymond~A. Yeh, and Greg Shakhnarovich.
\newblock Score jacobian chaining: Lifting pretrained 2d diffusion models for 3d generation.
\newblock In \emph{IEEE Conference on Computer Vision and Pattern Recognition (CVPR)}, 2023{\natexlab{a}}.

\bibitem[Wang et~al.(2018)Wang, Girshick, Gupta, and He]{wang2018non}
Xiaolong Wang, Ross Girshick, Abhinav Gupta, and Kaiming He.
\newblock Non-local neural networks.
\newblock In \emph{IEEE Conference on Computer Vision and Pattern Recognition (CVPR)}, 2018.

\bibitem[Wang et~al.(2023{\natexlab{b}})Wang, Lu, Wang, Bao, Li, Su, and Zhu]{wang2023prolificdreamer}
Zhengyi Wang, Cheng Lu, Yikai Wang, Fan Bao, Chongxuan Li, Hang Su, and Jun Zhu.
\newblock Prolificdreamer: High-fidelity and diverse text-to-3d generation with variational score distillation.
\newblock In \emph{Advances in Neural Information Processing Systems (NeurIPS)}, 2023{\natexlab{b}}.

\bibitem[Wu et~al.(2016)Wu, Zhang, Xue, Freeman, and Tenenbaum]{jiajun20163dgan}
Jiajun Wu, Chengkai Zhang, Tianfan Xue, William~T Freeman, and Joshua~B Tenenbaum.
\newblock Learning a probabilistic latent space of object shapes via 3d generative-adversarial modeling.
\newblock In \emph{Advances in Neural Information Processing Systems (NeurIPS)}, 2016.

\bibitem[Xie et~al.(2023)Xie, Bhatnagar, and Pons-Moll]{xie2023vistracker}
Xianghui Xie, Bharat~Lal Bhatnagar, and Gerard Pons-Moll.
\newblock Visibility aware human-object interaction tracking from single rgb camera.
\newblock In \emph{IEEE Conference on Computer Vision and Pattern Recognition (CVPR)}, 2023.

\bibitem[Xue et~al.(2023)Xue, Bhatnagar, Marin, Sarafianos, Xu, Pons-Moll, and Tung]{xue2023nsf}
Yuxuan Xue, Bharat~Lal Bhatnagar, Riccardo Marin, Nikolaos Sarafianos, Yuanlu Xu, Gerard Pons-Moll, and Tony Tung.
\newblock Nsf: Neural surface fields for human modeling from monocular depth.
\newblock In \emph{IEEE International Conference on Computer Vision (ICCV)}, 2023.

\bibitem[Youwang et~al.(2021)Youwang, Ji-Yeon, Joo, and Oh]{Youwang2021Unified3M}
Kim Youwang, Kim Ji-Yeon, Kyungdon Joo, and Tae-Hyun Oh.
\newblock Unified 3d mesh recovery of humans and animals by learning animal exercise.
\newblock In \emph{British Machine Vision Conference (BMVC)}, 2021.

\bibitem[Youwang et~al.(2022)Youwang, Ji-Yeon, and Oh]{youwang2022clipactor}
Kim Youwang, Kim Ji-Yeon, and Tae-Hyun Oh.
\newblock {CLIP}-{A}ctor: Text-driven recommendation and stylization for animating human meshes.
\newblock In \emph{European Conference on Computer Vision (ECCV)}, 2022.

\bibitem[Youwang et~al.(2023)Youwang, Hyun, Sung-Bin, Nam, Ju, and Oh]{youwang2023neuface}
Kim Youwang, Lee Hyun, Kim Sung-Bin, Suekyeong Nam, Janghoon Ju, and Tae-Hyun Oh.
\newblock A large-scale {3D} face mesh video dataset via neural re-parameterized optimization.
\newblock \emph{arXiv preprint, arXiv:2310.03205}, 2023.

\bibitem[Zielonka et~al.(2022)Zielonka, Bolkart, and Thies]{zielonka2022mica}
Wojciech Zielonka, Timo Bolkart, and Justus Thies.
\newblock Towards metrical reconstruction of human faces.
\newblock In \emph{European Conference on Computer Vision (ECCV)}, 2022.

\bibitem[Zuffi et~al.(2017)Zuffi, Kanazawa, Jacobs, and Black]{zuffi2017smal}
Silvia Zuffi, Angjoo Kanazawa, David Jacobs, and Michael~J. Black.
\newblock {3D} menagerie: Modeling the {3D} shape and pose of animals.
\newblock In \emph{IEEE Conference on Computer Vision and Pattern Recognition (CVPR)}, 2017.

\end{thebibliography}
}

\normalsize
\maketitlesupplementary
\newtheorem{assume}{Assumption}
\newtheorem{definition}{Definition}
\newtheorem{lemma}{Lemma}

\setcounter{section}{0}
\setcounter{figure}{0}
\setcounter{table}{0}
\setcounter{equation}{0}

\renewcommand{\thesection}{\Alph{section}}
\renewcommand{\thefigure}{S\arabic{figure}}
\renewcommand{\thetable}{S\arabic{table}}
\renewcommand{\theequation}{\alph{equation}}


In this supplementary material, we provide additional details and results that are not
included in the main paper due to the space limit.
The attached video includes a brief introduction and interesting qualitative results of $\ours$.


\section{Details of $\ours$}

\subsection{DC-PBR: Network Design}\label{sec:network}
The main contribution of our work is the proposed $\dcpbr$ parameterization for optimizing the physically-based rendering (PBR) texture maps. 
Instead of pixel-based parameterization of the PBR texture maps, we introduce the fixed random noise input $\bz\sim\mathcal{N}(0,\bI)\in\Real^{H\times W\times 3}$, and a randomly initialized U-Net with skip connections, $\mathcal{T}_\theta$.
We obtain the PBR texture map $[\bK^{\text{d}}_{\theta}, \bK^{\text{rm}}_{\theta}, \bK^{\text{n}}_{\theta}]=\mathcal{T}_\theta(\bz)\in\Real^{H\times W\times (3+2+3)}$, for every iteration of the synthesis optimization.

Our design choice of $\dcpbr$ is inspired by the Deep Image Prior~\cite{ulyanov2018dip}, and we extended it to re-parameterize the PBR texture maps for the text-driven texture map synthesis task.
We use an encoder-decoder (``hourglass'') architecture with skip connections between encoder and decoder features for our neural re-parameterization, $\dcpbr$ $\mathcal{T}_\theta$. 
For the network hyperparameters, we used the \emph{default architecture} of the Deep Image Prior, \ie, five levels of downsampling and upsampling layers for the encoder and decoder. 
We encourage readers to refer to the details in Fig. 21 of Deep Image Prior.
We empirically set the learning rate as $5\cdot10^{-4}$ and the total iteration for PBR texture synthesis as 1000.

\subsection{Details of SDS Loss}
Recall that we optimize the $\dcpbr$ given the text with the Score-Distillation Sampling (SDS).

\paragraph{SDS Loss for Multi-view Mesh Images}
We adopt some engineering to obtain high-fidelity and multi-view consistent PBR texture maps. 
When computing the SDS loss, we need the rendered image of the textured mesh. We randomly sample camera poses in multi-view and render $N$ view images. 
We sample the elevation angle as $\varphi_\text{elev}\sim\mathcal{U}(-\frac{\pi}{3},\frac{\pi}{3})$, and the azimuth angle as $\varphi_\text{azim}\sim\mathcal{U}(0,2\pi)$. We set $N=4$ for most cases.
When optimizing $\dcpbr$ for humans and animals, we increase the generation quality of the face regions by additionally rendering the face-focused images.
We translate the mesh so that the head can be the center of the world coordinate and render it with $\varphi_\text{elev}\sim\mathcal{U}(-\frac{\pi}{6},\frac{\pi}{3})$ and $\varphi_\text{azim}\sim\mathcal{U}(0,2\pi)$. For human and animal cases, we use a total $N=8$ views for computing SDS loss, where four views are for the full body, and the others are for face regions.
We also use the directional text prompt engineering 
as in prior arts~\cite{poole2022dreamfusion,chen2023fantasia} 
to mitigate the ``Janus problem''.

When computing the SDS loss, at each iteration, we synchronize the noise $\epsilon$ and the noising timestep $t$ for multi-view rendered images, \ie, we randomly sample a single $\epsilon$ and $t$ for each synthesis iteration and add the same amount of noise to the multi-view mesh images.
Finally, we sample the noising timestep $t$, from the distribution $\mathcal{U}(t_\text{min},t_\text{max})$. We start by $[t_\text{min},t_\text{max}]=[0.2, 0.98]$, and linearly narrows down the distribution so that it become $[t_\text{min},t_\text{max}]=[0.3, 0.5]$ by the end. We empirically set the ranges for $t_\text{min}$ and $t_\text{max}$.

\paragraph{Why is SDS Loss a Noisy Signal?}
In the main paper, we denote that the SDS loss is a noisy signal. 
By noisy, we refer to the incoherent nature of the SDS loss.
From Sec.~\colorref{3.1} and Eq.~\colorref{1} in the main paper, we notice the SDS loss is dependent on the randomly sampled Gaussian noise $\epsilon\sim\mathcal{N}(0,\bI)$ and the noising timestep $t\sim\mathcal{U}(t_\text{min},t_\text{max})$.
We sample $\epsilon$ and $t$ for every iteration of the PBR texture map synthesis; thus, the SDS loss is highly likely to give incoherent direction for updating the $\dcpbr$ $\mathcal{T}_\theta$. 
Moreover, as aforementioned, we use multi-view rendered images. 
Multi-view images contain different visible mesh parts, potentially providing incoherent update directions for $\mathcal{T}_\theta$.

\begin{figure}[t]
\centering
\includegraphics[width=\linewidth]{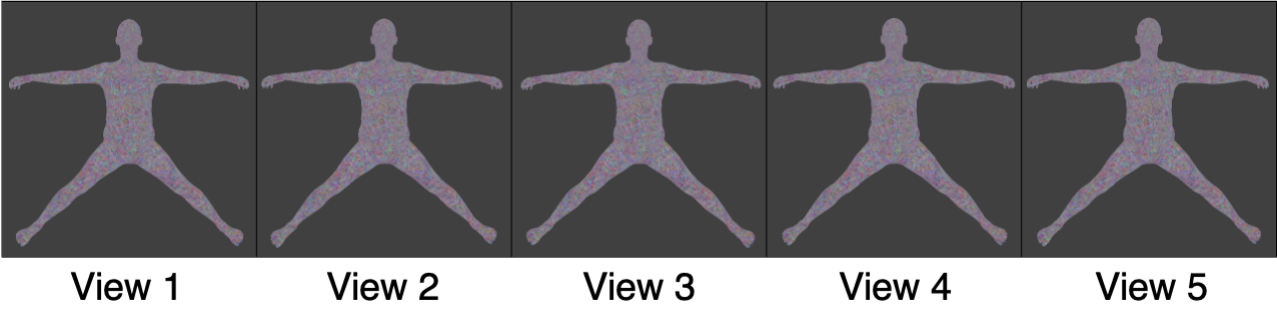}\vspace{-2mm}
   \caption{
    (For SDS analysis) Textured mesh image rendered from adjacent camera views $-2^{\circ}<\varphi_\text{elev}<2^{\circ}$ and $-2^{\circ}<\varphi_\text{azim}<2^{\circ}$. The images are slightly different but look almost identical.
   }
\label{fig:adj_view}
\end{figure}

To show the incoherent SDS gradient, we design a toy example. 
Given a mesh and a text input, we render the mesh in $N$ adjacent views, \ie, we sample elevation and azimuth in the range $-2^{\circ}<\varphi_\text{elev}<2^{\circ}$ and $-2^{\circ}<\varphi_\text{azim}<2^{\circ}$.
%
Since the camera views are closely distributed, the rendered mesh images would look almost identical (see \Fref{fig:adj_view}).

We investigated the backward gradients $\nabla\mathcal{L}_\text{SDS}$ applied on the diffuse map $\bK^{\text{d}}_\theta$, computed from each view, with the identical text prompt and $\epsilon$ and $t$.
%
We obtain a flattened, stacked gradient matrix from per-view SDS loss gradients on the diffuse map. Formally, we obtain the gradient matrix $\bG\in\Real^{N\times F}$, where $N$ denotes the number of rendered views, and $F$ denotes the flattened dimension of the gradient.
Even though we compute the SDS loss with the same text prompt, same $\epsilon$ and $t$, and the almost identical rendered images indistinguishable in eyes, we observe that the gradient matrix $\bG$ is in high rank. 
We first obtain the singular values of the gradient matrix $\bG$ using the Singular Value Decomposition (SVD) and investigate the ratio of all the singular values with the smallest singular value.
From a toy example, where $N=5$, we obtain the singular value ratio as $[2.1868, 1.7937, 1.7838, 1.6689, 1.0000]$, \ie, the highest and lowest singular values do not deviate too much in terms of the scale, showing $\bG$ is high rank.
In other words, the SDS loss, computed with a pre-trained text-to-image diffusion model, provides incoherent guidance from multiple views and guides the optimization in incoherent directions. 

We claim the SDS loss is a noisy signal from this observation. In Sec.~\colorref{4.2}, we empirically show our $\dcpbr$ filters out the high-frequency noisy signal by inherently scheduling the optimization curriculum. Thus, $\dcpbr$ is effective when combined with the noisy SDS loss. 

\subsection{Details of Frequency-based Analysis}
\begin{figure}[t]
\centering
\includegraphics[width=\linewidth]{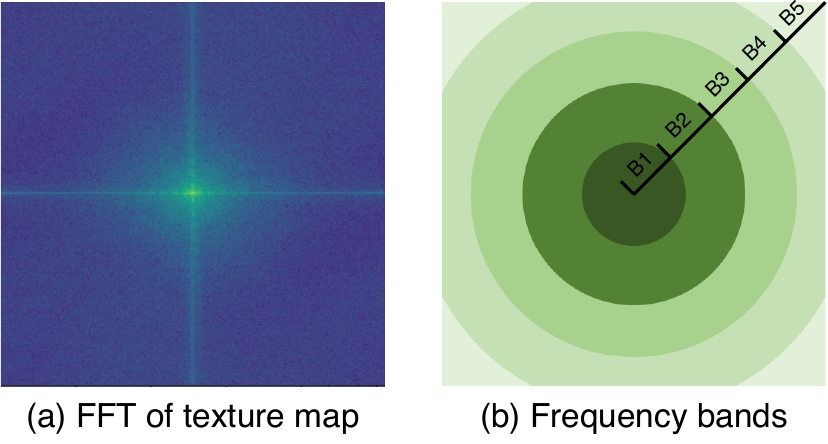}\vspace{-2mm}
   \caption{(a) Visualization of the FFT image of the diffuse map. (b) Visualization of the pre-defined non-overlapping frequency bands. Center: low frequency, Outer: high frequency.
   }
\label{fig:fft_image}
\end{figure}

We plot the energy-iteration plot in our frequency-based analysis of the proposed $\dcpbr$ (Fig.~\colorref{4} in the main paper).
We first performed the Fast Fourier Transform (FFT) of the diffuse texture map obtained in each iteration.
See \Fref{fig:fft_image}\colorref{a} for the FFT result.
Then, we define five non-overlapping frequency bands, depending on the radius from the center of the FFT image, as in \Fref{fig:fft_image}\colorref{b}. Note that the ranges of the frequency bands are fixed during the optimization.
Finally, we compute the energy of each frequency band by summing all the frequency response magnitudes (either absolute value or square works) in each frequency band.

In Fig.~\colorref{4}-left, the optimization starts from a monotonic gray image,
which has high zero-frequency energies, \ie, DC component. 
%
However, note that we visualize each frequency band’s `average' energy.
DC component 
occupies only a tiny area
($\ll1\%$) in the lowest frequency band; 
thus, the pixel brightness hardly affects the band's energy.

During $\ours$ optimization, our $\dcpbr$ representation automatically schedules the curriculum to learn low frequency first, then mid frequency, and filters out the highest frequency, \ie, noise.

\begin{figure}[t]
\centering
\includegraphics[width=\linewidth]{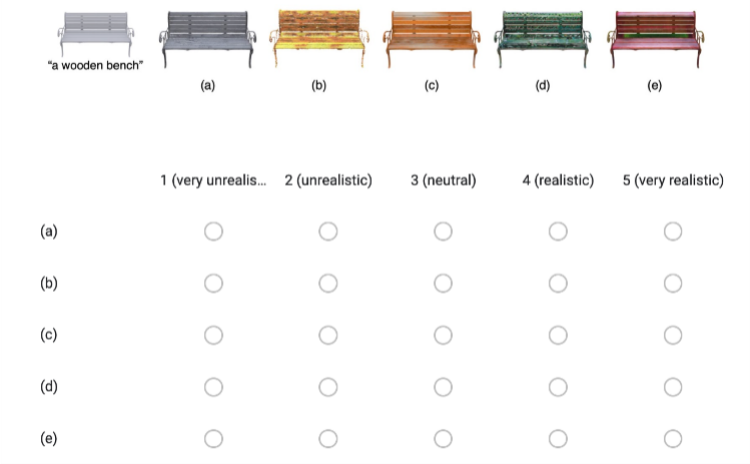}
   \caption{The users were asked to rate the realism of the rendered mesh images, textured with five different methods. The order of the methods was randomly shuffled for each question.
   }
\label{fig:user_interface}
\end{figure}

\begin{figure}[t]
\centering
\includegraphics[width=\linewidth]{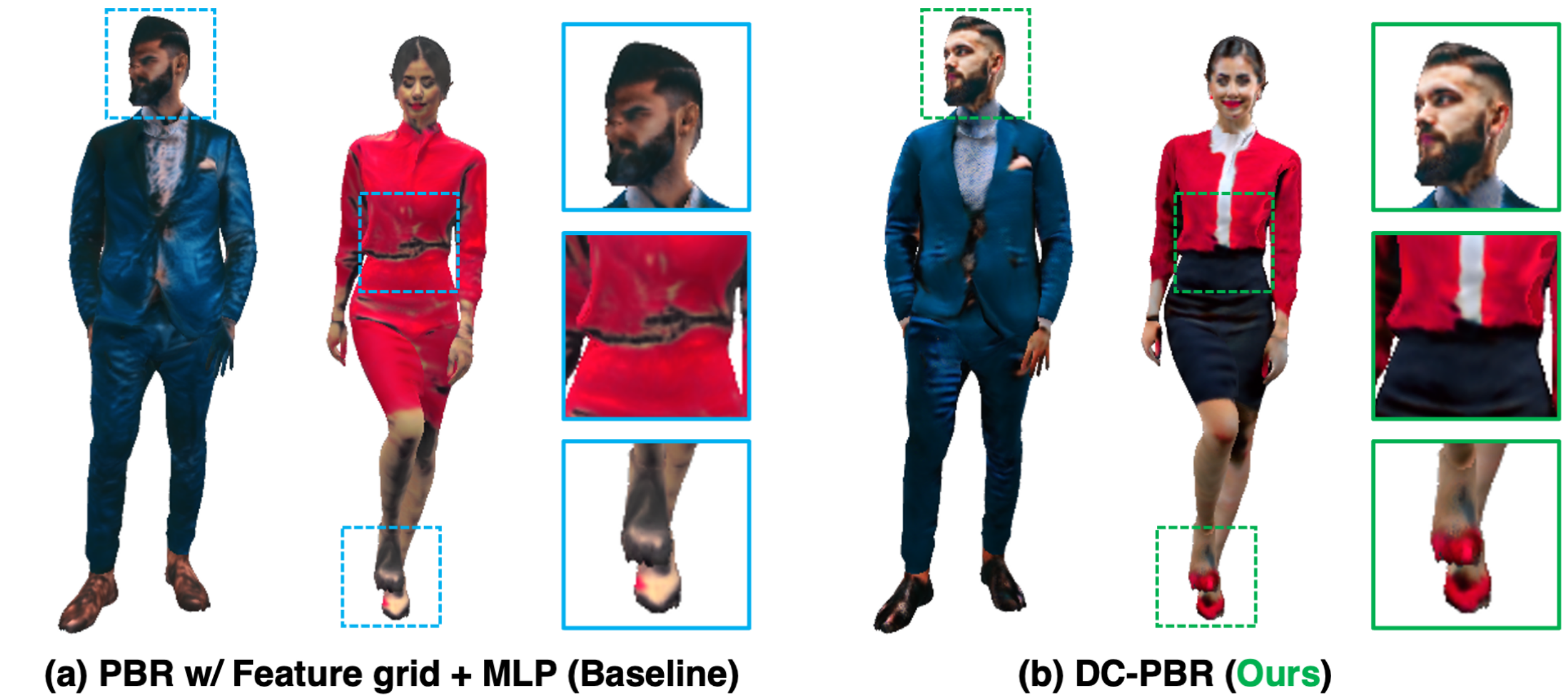}
   \caption{
   {Feature grid+MLP (baseline) vs. DC-PBR (ours)}
   }
   \vspace{-3.5mm}
\label{fig:reb_ingp}
\end{figure}

\begin{figure*}[t]
\centering
\includegraphics[width=1.0\linewidth]{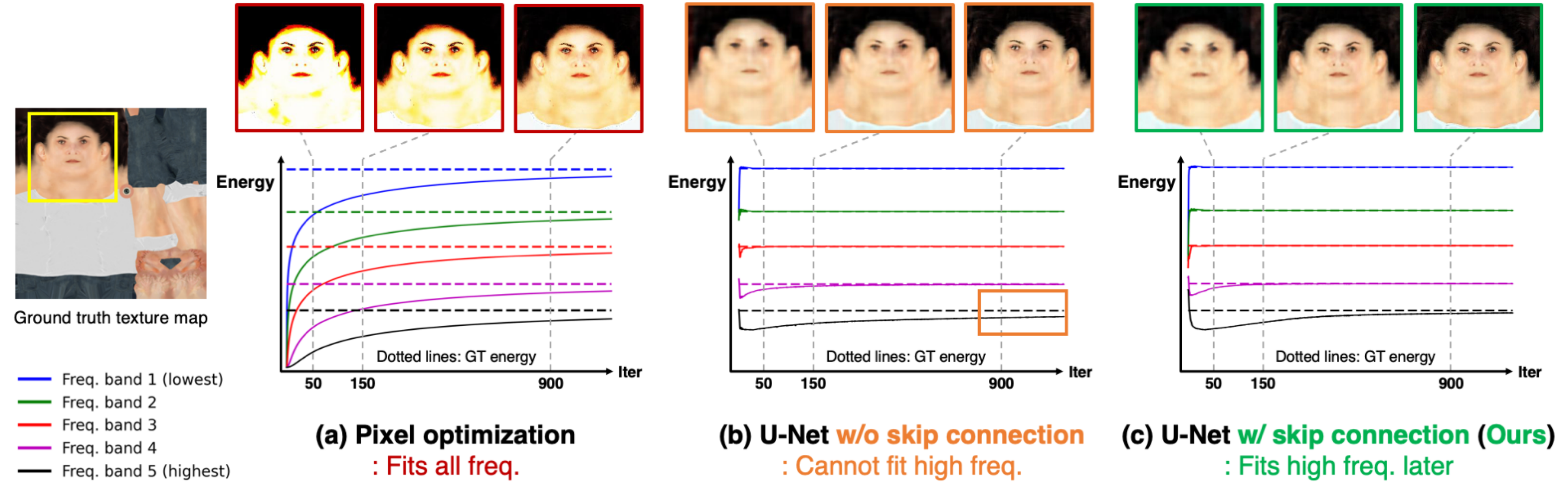}
   \caption{
   \textbf{Effect of skip connection: Texture map fitting.}
   When fitting a texture map with different parameterizations, U-Net without skip connections fails to fit the highest frequency band. This hints that the skip connections are responsible for representing fine-grained details. 
   }
\label{fig:supp_unet_fit}
\end{figure*}

\begin{figure*}[t]
\centering
\includegraphics[width=\linewidth]{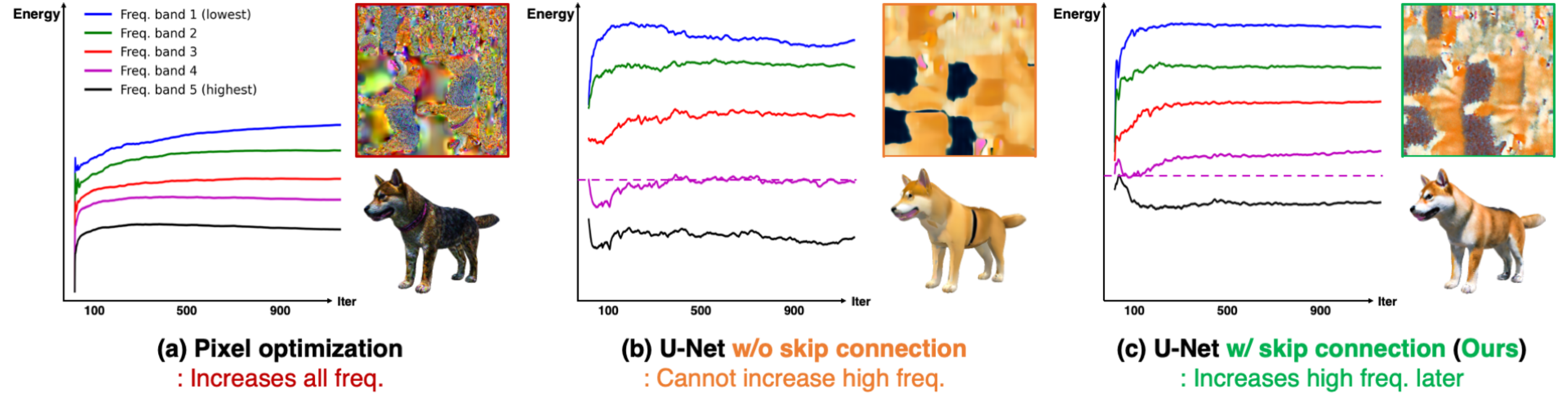}
   \caption{
   \textbf{Effect of skip connection: Text-driven texture synthesis.} 
   When synthesizing PBR texture maps with different parameterizations, U-Net without skip connections cannot increase the mid-to-high frequency band (purple). 
   Proposed DC-PBR with U-Net and skip connections can synthesize fine-grained details in texture maps, resulting in high-fidelity synthesis results.
   }
\label{fig:supp_unet_syn}
\end{figure*}

\subsection{Details of User Study}
We conducted a user study to assess the realism of the synthesized PBR texture maps. 
We showed ten untextured meshes from the Objaverse~\cite{objaverse} dataset and showed five different results obtained from the methods: Latent-Paint~\cite{metzer2022latent}, Fantasia3D~\cite{chen2023fantasia}, Text2Tex~\cite{chen2023text2tex}, TEXTure~\cite{richardson2023texture}, and $\ours$ (ours).
We asked 30 users with engineering/non-engineering backgrounds to rate the realism of the rendered images in the score range 1 to 5, \ie, 1: very unrealistic, 2: unrealistic, 3: neutral, 4: realistic, and 5: very realistic.
The order of the methods was randomly shuffled for fairness.
The interface of the user study is shown in \Fref{fig:user_interface}.
%

\subsection{Details \& Discussion about Optimization Time}
The optimization time for synthesizing PBR texture maps takes about 15 min. for general object meshes in Objaverse~\cite{objaverse}. 
For more complex cases, such as 3D humans or animals, we additionally use face-focused mesh renderings; thus, it takes about 30 min. to complete.
We optimized both the baseline (Eq.~(4) in the main paper, pixel optimization) and our DC-PBR until convergence, and there's no significant time difference.
%
Instead, DC-PBR helps the optimization with noisy SDS loss find a better solution than 
the vanilla pixel optimization.
We use a single NVIDIA RTX A6000 GPU for the optimization.

Extending $\ours$ to large-scale 3D scenes would be an interesting future direction. 
However, $\ours$'s PBR texture generation for large-scale scenes would take a longer optimization time. 
As suggested in FPRF~\cite{kim2024fprf}, we may try semantic feature distillation to accelerate the optimization when stylizing large-scale 3D scenes using SDS.

\section{Additional Experiment}

\subsection{More Baseline for PBR Representation}
For a PBR representation baseline, other than direct pixel optimization (Eq.~(4) in the main paper), we implement a 
multi-resolution hash encoding of grid features and subsequent MLP~\cite{mueller2022ingp} to model
the disentangled PBR texture maps. 
%
%
In \Fref{fig:reb_ingp}, our DC-PBR representation yields smoother and more vivid texture results than the new baseline.
In \Tref{tab:reb_fid}, our
DC-PBR 
obtains a better FID score than 
other PBR representation baselines (Pixel optim., \& Feat. grid+MLP).
We postulate that the SDS gradients only propagate to local hash grids in the new baseline, lacking non-local texture smoothness, as supported in~\cite{li2023neuralangelo}.
On the other hand, the spatial CNN kernels of our DC-PBR are beneficial in naturally imposing texture smoothness.

\begin{figure*}[t!]
\centering
\includegraphics[width=0.96\linewidth]{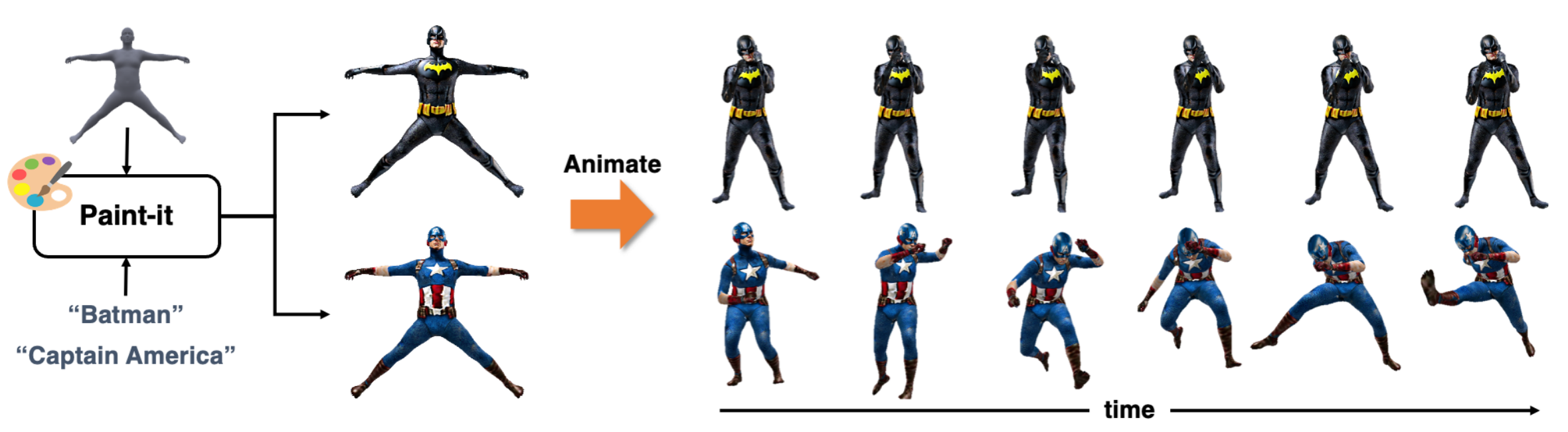}\vspace{-1.5mm}
   \caption{
   \textbf{$\ours$: Dynamic virtual 3D humans.}  
   We synthesize PBR texture maps given text and rigged mesh, \eg, SMPL-X~\cite{pavlakos2019smplx}, using $\ours$. 
   %
   Then, we animate the textured 3D humans using sequential pose parameters (can be retrieved~\cite{youwang2022clipactor} or generated~\cite{kim2022flame}).
   }\vspace{-1.5mm}
\label{fig:supp_motion}
\end{figure*}

\subsection{Effects of DC-PBR Design Choice}
In this section, we investigate the effect of our DC-PBR design choice. Specifically, we study the effect of our U-Net with skip connections in synthesizing the PBR texture maps.

As discussed in the main paper and \Sref{sec:network}, we use a \emph{randomly initialized} U-Net with skip connections, shortly, U-Net+skip. 
Deep Image Prior~\cite{ulyanov2018dip} used U-Net+skip and claimed skip connections inherently promote self-similarity across multi-scales, which is beneficial for inverse problems.
We wanted to investigate how the skip connections affect the DC-PBR optimization with SDS loss. Thus, we conduct the same experiment as in Sec.~\colorref{4} of the main paper, but with U-Net, without (\emph{w/o}) skip connections.

\paragraph{Fitting behavior}
Following the experiment in Sec.~\colorref{4.1}, we fit a randomly initialized U-Net \emph{w/o} skip connections given a ground-truth texture map.
In \Fref{fig:supp_unet_fit}\colorref{b}, 
the energy-iteration plot shows that parameterizing a texture map with U-Net \emph{w/o} skip connection fails to fit high frequency.

\paragraph{DC-PBR synthesis behavior}
Similarly, for our task, \ie, text-driven DC-PBR optimization, U-Net \emph{w/o} skip connection fails to increase the mid-to-high frequency band (purple line), resulting in blurry texture maps. In contrast, our DC-PBR, parameterized in U-Net+skip, successfully increases the mid-to-high frequency band, synthesizing fine-grained texture maps.
We conclude that the skip connections are in charge of synthesizing fine-grained, high-frequency details of texture maps. 
This observation aligns with the Deep Image Prior's claim, where skip connections benefit the inverse problems with multi-scale feature awareness.
We additionally showed the frequency level behavior of the skip connection through the experiments (see \Fref{fig:supp_unet_syn}).

\begin{table}
    \centering
    \resizebox{\linewidth}{!}{
    \begin{tabular}{lcccc}
        \toprule
             & Pixel optim. & Only $\bK^\text{d}$ & Feat. grid+MLP & \textbf{DC-PBR (Ours)} \\
         \cmidrule{1-5}
            FID ($\downarrow$) & 216.6 & 59.39 & 55.98 & \textbf{34.46} \\
        \bottomrule
    \end{tabular}
    }
    \caption{{FID for ablation study}
    }
    \vspace{-3.5mm}
    \label{tab:reb_fid}
\end{table}

\begin{figure*}[htbp]
    \centering
       \includegraphics[width=0.925\linewidth]{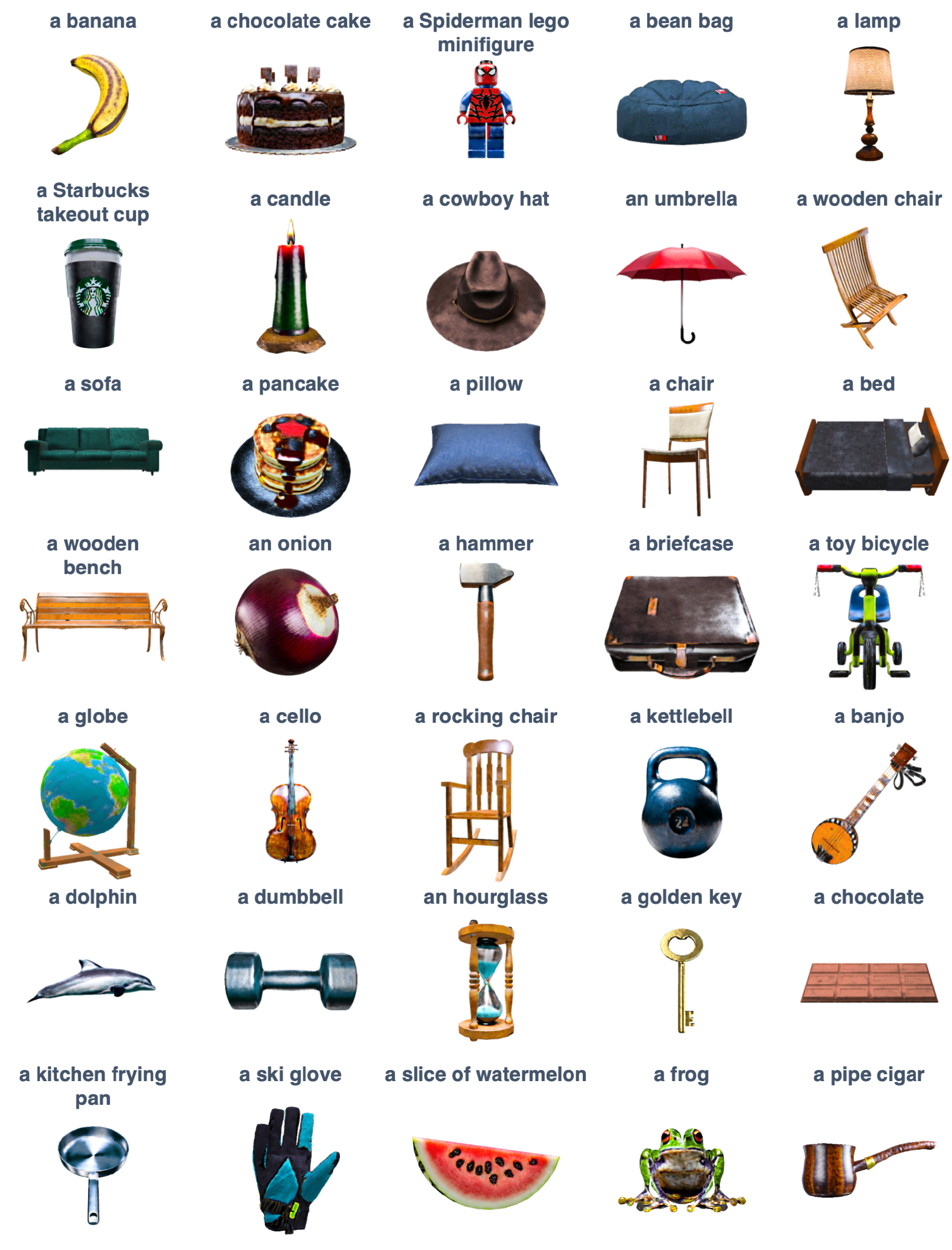}
       \caption{\textbf{Qualitative results of $\ours$: Objaverse dataset~\cite{objaverse}}. 
       Given any untextured mesh from the existing mesh database, $\ours$ synthesizes high-fidelity, locally smooth, and realistic object PBR texture maps.
       }   
    \label{fig:supp_objaverse} 

\end{figure*}

\begin{figure*}[htbp]
    \centering
       \includegraphics[width=0.9\linewidth]{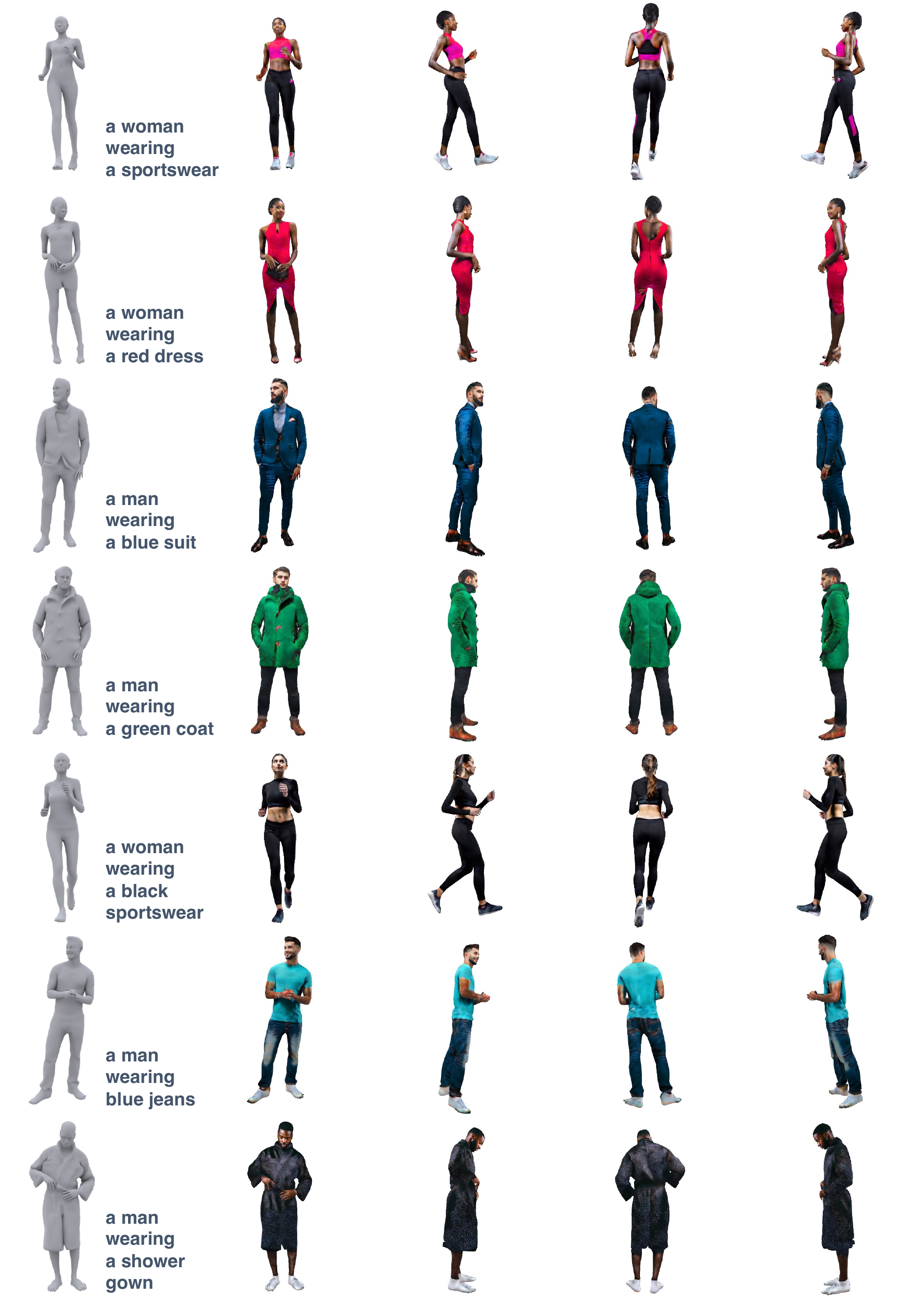}
       \caption{\textbf{Qualitative results of $\ours$: RenderPeople dataset~\cite{renderpeople}}. Given untextured clothed human meshes, $\ours$ synthesizes high-fidelity, vivid, and multi-view consistent human and cloth PBR textures. We render four different views of the textured mesh.
       }   
    \label{fig:supp_renderpeople} 

\end{figure*}

\begin{figure*}[htbp]
    \centering
       \includegraphics[width=0.9\linewidth]{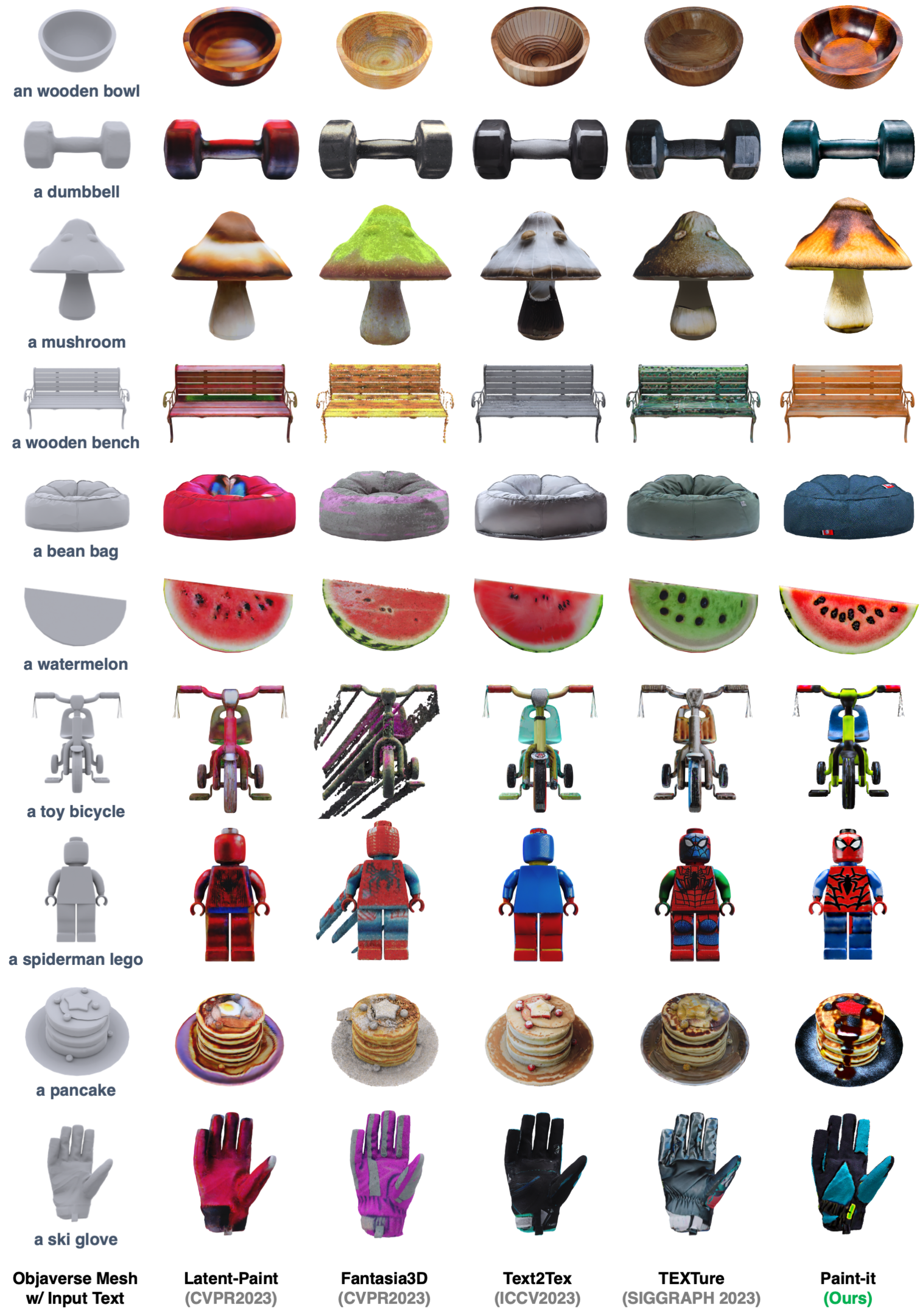}
       \caption{\textbf{Comparison results: Objaverse dataset~\cite{objaverse}}.
       %
       }   
    \label{fig:supp_comp_obja} 

\end{figure*}

\begin{figure*}[htbp]
    \centering
       \includegraphics[width=0.9\linewidth]{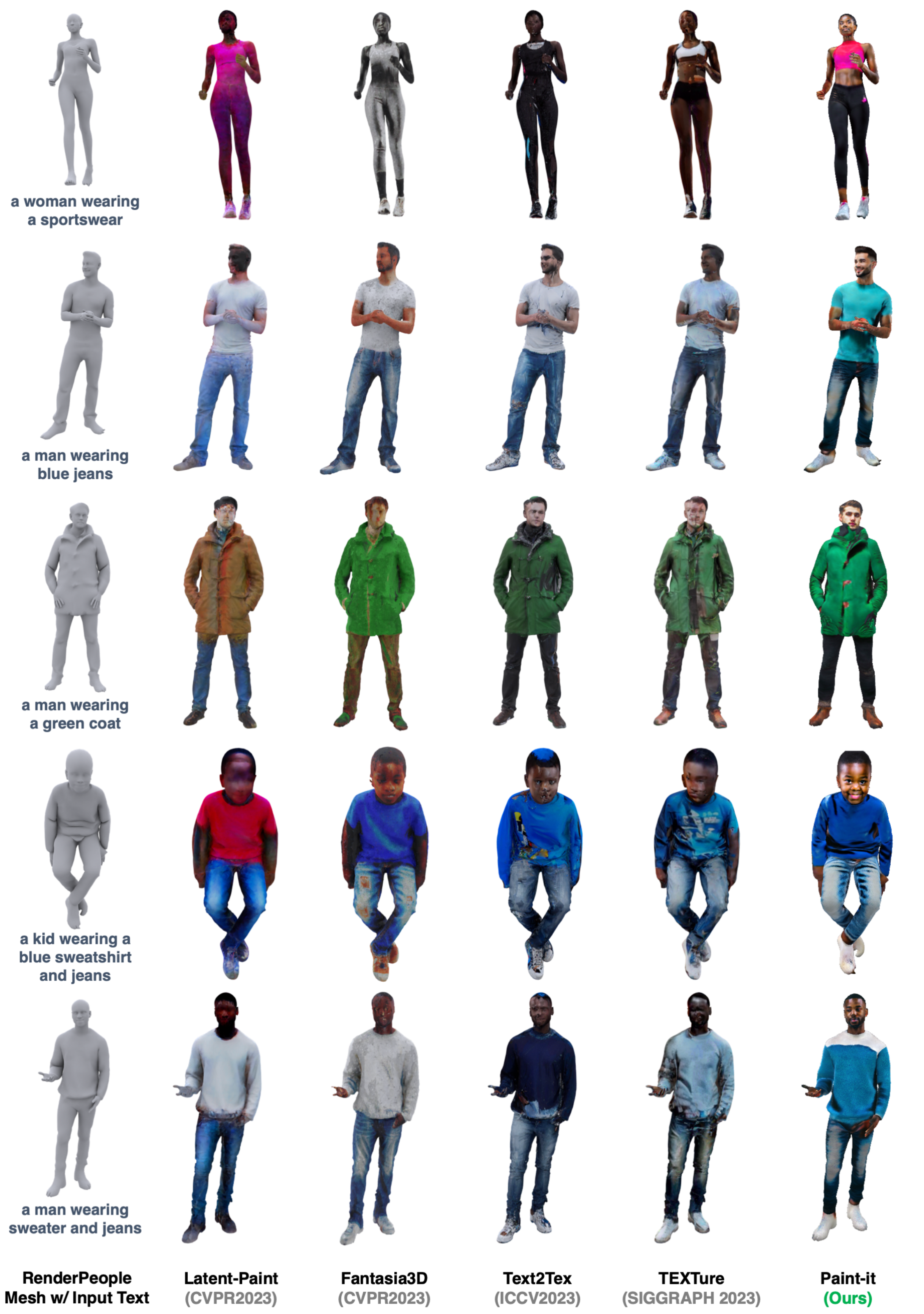}
       \caption{\textbf{Comparison results: RenderPeople dataset~\cite{renderpeople}}. 
       %
       }   
    \label{fig:supp_comp_rp} 

\end{figure*}

\section{Additional Results}

\subsection{More Quantitative Comparison}
In Tab.~\ref{tab:reb_fid}, we report FID scores for the ablation study (in Sec.~\colorref{5.3} in the main paper). 
We used 410 meshes from Objaverse to get the real sample set and added 50 meshes for the generated sample set.
The results support that our full DC-PBR 
enhances the realism of the generated texture.

\subsection{More Qualitative Results}
We provide more qualitative results of our $\ours$.
Given untextured meshes from Objaverse~\cite{objaverse} and RenderPeople~\cite{renderpeople}, we obtain the text prompts from
1) manually writing the requirements, \eg, a Spiderman lego minifigure, or 2) generating an automatic text caption using multi-modal large-language models, \eg, GPT-4. 
Then, we conduct $\ours$ optimization to synthesize PBR texture maps and render the textured meshes (see \Fref{fig:supp_objaverse} and \Fref{fig:supp_renderpeople}).

\subsection{More Comparison Results}
In Figs.~\ref{fig:supp_comp_obja} and \ref{fig:supp_comp_rp}, we provide more qualitative results that compare $\ours$ and recent competing methods~\cite{metzer2022latent,chen2023fantasia,chen2023text2tex,richardson2023texture}.
As in Fig.~\colorref{6} of the main paper, we synthesize texture maps using each method for the same untextured meshes and text prompts.
Overall, $\ours$ synthesizes much realistic and vivid texture on the meshes, thanks to the PBR texture representation and texture smoothness induced by our proposed DC-PBR.
Note that Fantasia3D~\cite{chen2023fantasia} also synthesizes PBR materials, but in a per-point independent manner; thus, it lacks texture smoothness and yields substantial jitterings.
Moreover, given an untextured mesh, Fantasia3D converts the mesh into a signed distance function (SDF) representation, DMTet~\cite{shen2021dmtet}. 
Such auxiliary re-meshing optimization leads to severe geometric quality degradation, \eg, floating artifacts on a toy bicycle example, in \Fref{fig:supp_comp_obja}.

\subsection{\textbf{$\ours$} for Animated Meshes}
$\ours$ can also synthesize high-quality PBR texture maps for animatable meshes and generate dynamic 3D assets.
Since $\ours$ does not perform a re-meshing process and preserves the original UV texture coordinates, we can synthesize maps for any rigged meshes, \eg, T-posed human mesh, and animate with any motion sequences.

In this paper, we used SMPL-X~\cite{pavlakos2019smplx}. We first take the canonical posed SMPL-X mesh and synthesize PBR texture maps using $\ours$. 
To animate the textured avatars,
one may use the motion captured mesh sequences of 3D human bodies~\cite{cho2022FastMETRO,goel2023humans,xue2023nsf,xie2023vistracker,vladimir2021hps}, faces~\cite{zielonka2022mica,youwang2023neuface,feng2021deca} or even animals~\cite{Youwang2021Unified3M,biggs2020wldo}.
Generative models for natural body or facial motions~\cite{rempe2021humor,tevet2023human,sungbin2024laughtalk} could also be applied for animation.
We may also use the posed meshes and perform advanced augmentations as proposed in CLIP-Actor~\cite{youwang2022clipactor}.
We visualize the synthesized animated meshes in \Fref{fig:supp_motion}.

\end{document}